\documentclass[journal]{IEEEtran}
\usepackage{pifont}
\ifCLASSINFOpdf
\else
\fi
\usepackage{latexsym}
\usepackage{todonotes}
\usepackage{multirow}
\usepackage{subcaption}
\usepackage{microtype}
\usepackage{ccicons}
\usepackage[pdflang={en-US},pdftex]{hyperref}
\usepackage{color}
\usepackage{booktabs}
\usepackage{textcomp}
\usepackage{amsmath}
\usepackage{amssymb}

\usepackage{cleveref}

\newcommand{\cmark}{\ding{51}}%
\newcommand{\xmark}{\ding{55}}%

\begin{document}
%
\title{Text-based depression detection on sparse data}

%
%
%

\author{Heinrich~Dinkel,~\IEEEmembership{Student Member,~IEEE,}
        Mengyue~Wu,~\IEEEmembership{Member,~IEEE,}
        and~Kai~Yu,~\IEEEmembership{Senior Member,~IEEE}
\thanks{Mengyue Wu and Kai Yu are the corresponding authors.}}
\maketitle

\begin{abstract}
Previous text-based depression detection is commonly based on large user-generated data. Sparse scenarios like clinical conversations are less investigated. 
This work proposes a text-based multi-task BGRU network with pretrained word embeddings to model patients' responses during clinical interviews.
Our main approach uses a novel multi-task loss function, aiming at modeling both depression severity and binary health state.
We independently investigate word- and sentence-level word-embeddings as well as the use of large-data pretraining for depression detection.
To strengthen our findings, we report mean-averaged results for a multitude of independent runs on sparse data.
First, we show that pretraining is helpful for word-level text-based depression detection.
Second, our results demonstrate that sentence-level word-embeddings should be mostly preferred over word-level ones.
While the choice of pooling function is less crucial, mean and attention pooling should be preferred over last-timestep pooling.
Our method outputs depression presence results as well as predicted severity score, culminating a macro F1 score of 0.84 and MAE of 3.48 on the DAIC-WOZ development set.
\end{abstract}


%
\IEEEpeerreviewmaketitle

\begin{figure*}[htbp]
    \centering
    \begin{subfigure}[t]{0.5\textwidth}
        \centering
        \includegraphics[width=0.99\linewidth]{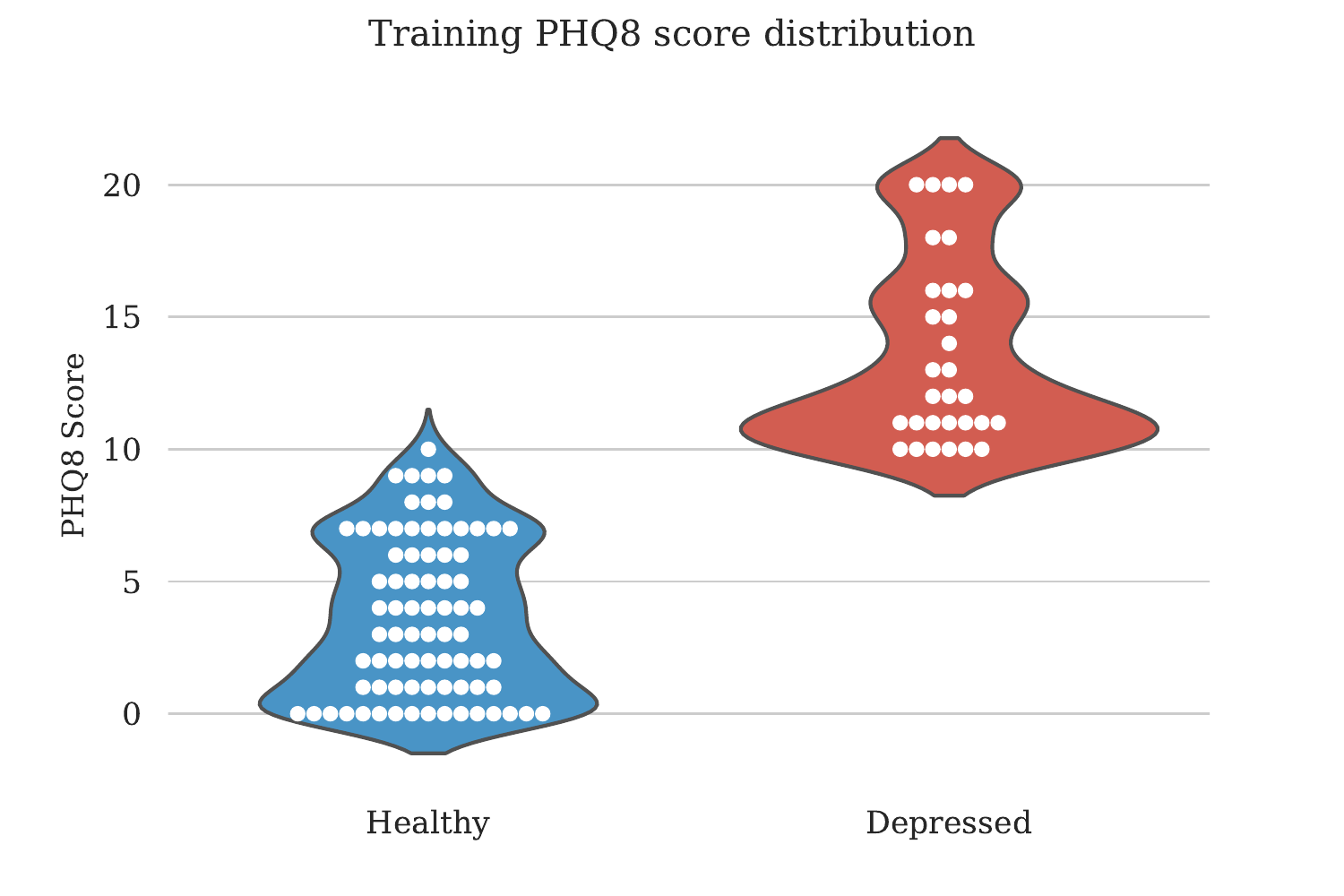}
    \end{subfigure}%
    \begin{subfigure}[t]{0.5\textwidth}
        \centering
        \includegraphics[width=0.99\linewidth]{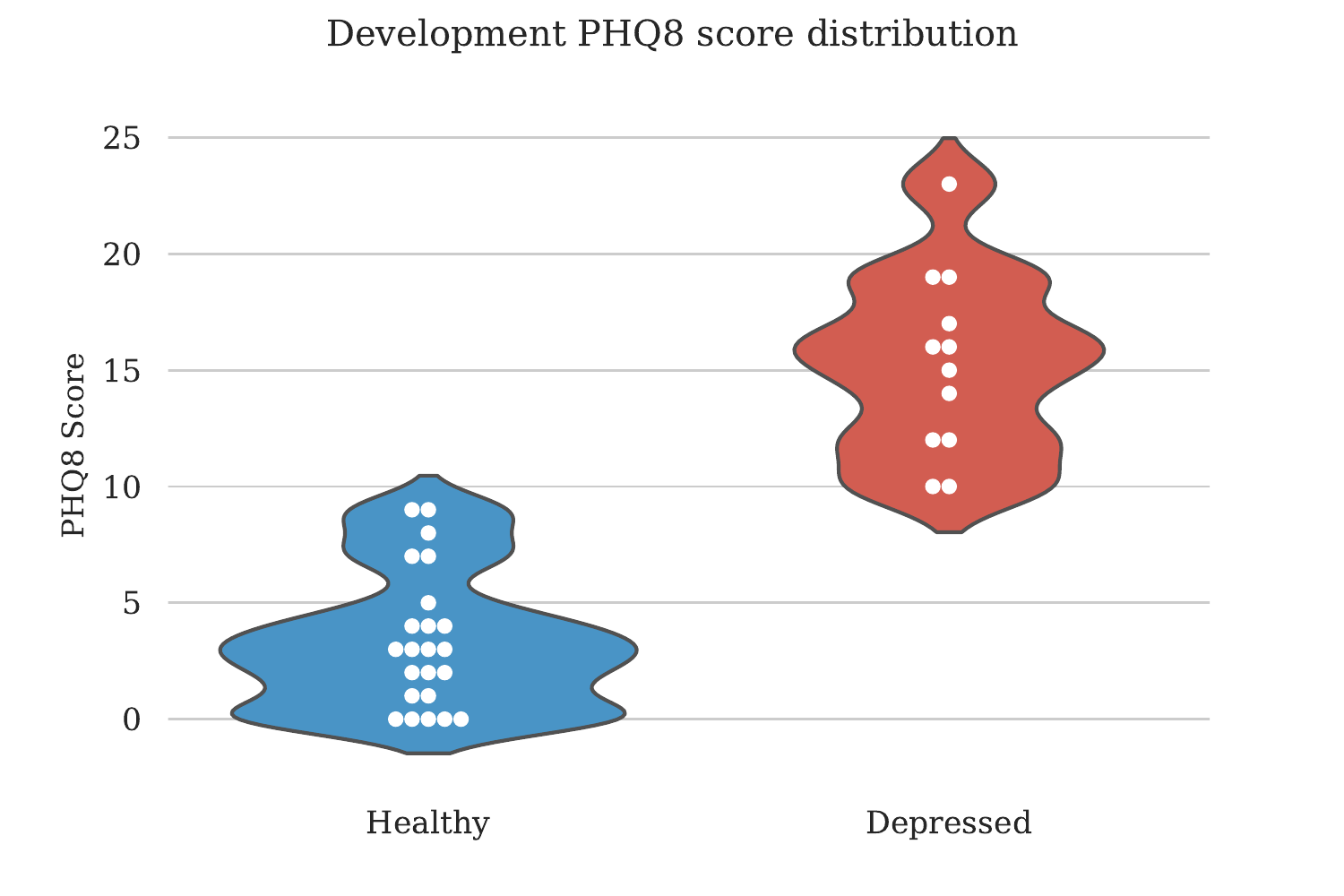}
    \end{subfigure}%
    \caption{DAIC-WOZ training and development data PHQ-8 score distribution. Each dot represents a respective patients' score.}
    \label{fig:data_distribution}
\end{figure*}

\section{Introduction}
\IEEEPARstart{D}{epression} is an illness that affects, knowingly or unknowingly, millions of people worldwide. Efficient and effective automatic depression diagnosis can be of substantial benefit. 
However, this is an arduous task since a variety of complicated symptoms are reported, and subjective clinical interview is the golden standard. Classic depression detection at its core is a binary classification problem, with classifiers explored from traditional methods like SVM~\cite{cohn2009detecting}, naive Bayes~\cite{deshpande2017depression}, decision tree~\cite{yang2016decision} and neural networks like long short term memory (LSTM)~\cite{su2018lstm} and convolutional neural network (CNN)~\cite{orabi2018deep}. 
Severity prediction can either be seen as a multi-class classification or a regression problem, usually associated with a psychological questionnaire score like Patient Health Questionnaire (PHQ)-8/9~\cite{zich1990screening} or Beck's Depression Inventory (BDI)~\cite{gilbody2007screening}. Though various deep learning models have been utilized (\cite{AlHanai2018,Haque2018, Williamson:2016:DDU:2988257.2988263,Zhao2020,bjorn2019,Rejaibi2020TowardsRD,Ma2016}), the assessment precision is far from satisfaction.
For instance, compared to regression modeling on other topics, the mean absolute errors and root mean squared errors reported in severity predictive tasks are quite high~\cite{Haque2018}. This again emphasizes the complexity of depression symptoms and the difficulty of precise predictions. 

In particular, text-based depression detection has been broadly investigated on user-generated data, e.g., a task in the CLEF eRisk challenge aims at depression severity prediction from data collected online, including questionnaire answering~\cite{losada2020erisk} and written social media texts~\cite{losada2017erisk}.
Similarly, CLPsych organized tasks for PTSD and anxiety detection from user-generated texts \cite{coppersmith2015clpsych}.
Initial studies using text-based approaches for early depression detection on the eRisk dataset have shown promising performance~\cite{Trotzek_metadata_for_early_detection, Trotzek_2018_word_embeddings}.
Different text feature sets have been explored, ranging from hand-crafted feature types such as n-grams, Bag of Words, Linguistic Inquiry and Word Count (LIWC)~\cite{Trotzek_2018_word_embeddings}, Paragraph Vector etc., to neural word embeddings like Word2Vec~\cite{mikolov2013efficient} which consists of the Continuous Bag of Words (CBoW) and the Skip-gram models, fastText~\cite{joulin2016bag,bojanowski2017enriching}, as well as GloVe~\cite{pennington2014glove}. 
The advantage of user-driven datasets is that they can generally lead to larger training corpora, where deep-learning methods have seen big success in recent years~\cite{Kong2020,Kong2019a,Chen2020,Hershey2017}. 

This study situates itself from a different angle: how to detect depression via text in sparse data scenarios.
Mainly, we are interested in conversational data between an interviewee and a clinical therapist.
It should be noted that self-generated data concerns a large number of users with potentially big data, while clinical conversations conducted for depression detection can be very limited. Clinical depression detection has been explored by a different challenge, namely Audio/Visual Emotion Challenge (AVEC) ~\cite{Ringeval:2017:ARD:3133944.3133953}, on a flagship dataset Distress Analysis Interview Corpus - Wizard of Oz (DAIC-WOZ)~\cite{Gratch_thedistress,DeVault:2014:SKV:2615731.2617415}, which includes video, audio along with its text transcripts. 
This published dataset only includes 107 participants; thus, the training scheme, feature selection, and pooling methods greatly differ from those in a large data setting. However, conversational data can potentially reveal more information on the participant's linguistic ability and cognitive function, therefore can provide a different angle towards depression detection. Hence, we would like to investigate robust depression 
prediction based on conversational text, with much less data compared to previous text sources. 

\subsection{Contribution}
This paper mainly aims at robust depression detection on sparse data: we first examine text embeddings at the word- and sentence- level then compare embeddings with and without pretraining; Second, we analyze different pooling methods that match the embeddings. 
Lastly, we adopt 5-Fold cross-validation in all our experiments.
Our results are reported on average as well as best performances. 

Accordingly, our main contributions include:
\begin{itemize}
    \item An innovative multi-task model design, combining binary depression detection with severity prediction.
    \item Investigating the usage of pretrained word/sentence embeddings to alleviate sparse-data depression detection problems.
    \item Investigating the performance difference between word-level and sentence-level embedding
    \item Providing analyses on different pooling functions that best match each respective front-end text-embedding.
\end{itemize}

The rest of the paper is organized as follows: an overview of the task and its related work is firstly provided in \Cref{sec:background}, followed by a detailed layout of our multi-task sequence modeling approach for word/sentence-level depression detection in \Cref{sec:approac}.
Then in \Cref{sec:experiments}, we introduce the utilized dataset and specify our training and evaluation framework. 
The proposed approach is then evaluated, and results are displayed in \Cref{sec:results}.
Conclusions and future work are drawn in \Cref{sec:conclusion}.

\section{Background and Related Work}
\label{sec:background}
\subsection{Conversational Depression Datasets}

To date, a number of research groups and hospitals are dedicated to publishing better quality and larger quantities of depression datasets for a review see~\cite{cummins2015review}.
However, publicly available conversational datasets appropriate for incorporating machine learning methods are surprisingly limited. 
The one most broadly used is DAIC-WOZ~\cite{Gratch_thedistress,DeVault:2014:SKV:2615731.2617415}, which encompasses 50 hours of data collected from 189 clinical interviews from a total of 142 patients. 
Two labels are provided for each participant: a binary diagnosis of depressed/healthy and the patient's eight-item \textbf{P}atient \textbf{H}ealth \textbf{Q}uestionnaire score (PHQ-8) metric~\cite{Kroenke2009}.
Low PHQ-8 scores represent a healthy patient, while high PHQ-8 scores represent possible depression symptoms.
Thirty speakers within the training (28 \%) and 12 within the development (34 \%) set are classified to have depression (binary value is set to 1). 
This database was previously used for the AVEC 2017~\cite{Ringeval:2017:ARD:3133944.3133953} challenge, stating that scores larger than $10$ are considered to be depressed. 
However, in reality, any questionnaire could merely work as a reference, and a clinical interview is the golden standard~\cite{martin2017depression}.

We analyzed the dataset to understand the difficulties involved when modeling this task. Overall, three observations can be made about depression detection from a data perspective: 1) the dataset itself is relatively insufficient; 2) the PHQ-8 distribution of training and development sets are quite different, ie., some scores are only seen in the training data; 3) the depression state and PHQ-8 score are correlated, but one characteristic does not necessarily predict the other. 

\subsection{Depression Detection Methods}
Automatic depression detection research can either predict the classification results or a severity score, to associate with the mental state label and PHQ-8 score. 
The depression presence is a binary classification model, predicting the healthy/depressed state of a given speaker. F1 score, precision, and recall are presented as the performance results. 
Different classification methods have been tested out over the last few decades with different feature extraction modifications. 
The severity prediction task is usually regressed with questionnaire scores, which is assessed by the mean absolute error (MAE) and root mean square error (RMSE). 
Classification and regression are traditionally seen as substitutions of each other. 
However, though they are closely correlated, there is still a distinction between the two. 
Thus, our work firstly adopted a multi-task training method to model classification and regression tasks at the same time, details of which will be shown in \Cref{sec:approac}.

\subsection{Text-based detection}

Many text-based depression detection studies stem from social media content, combined with its accompanying picture~\cite{gui2019cooperative} or by itself~\cite{Trotzek_2018_word_embeddings,Trotzek_metadata_for_early_detection,Kong2019a}. 
Conversational text generated from clinical interviews is rarely examined on its own, commonly investigated as semantic features or linguistic features along with other behavioral features \cite{AlHanai2018}.
\cite{Williamson:2016:DDU:2988257.2988263} exploited word representations with Global Vectors for Word Representation (GloVe), following a high-level feature learning method. 
Since a clinical conversation involves questions and answers from different parties, they separated semantic information into content and context analysis. 
It concluded that the semantic analysis of dialogue scripts via text-based features is the most promising depression detection method compared with other modalities.
\cite{AlHanai2018} compared different ways of modeling a conversation with the combination of audio and text features. 
In specific, they compared context-dependent, context-free, and sequence modeling methods using Word2Vec as their word embeddings. 
A recent study~\cite{bjorn2019} employed a hierarchical attention mechanism to model textual information at both word and sentence levels, aiming to connect word-level representations with sentence-level ones. 
The work used a GRU network and GloVe as its textual embeddings, outputting the binary prediction probability as its results.

As previously noted, clinical conversational depression datasets are sparse. Thus word/sentence embeddings trained from scratch might not be able to understand the context and represent the word/sentence effectively. 
At the same time, clinical interviews for depression detection are mostly consisted of regular conversations, with seldom use of medical-specific terms. Recently, general-purpose text-embeddings such as ELMo~\cite{Peters:2018} and BERT~\cite{devlin2018bert}, become popular due to their performance on many NLP benchmarks.
Compared to GloVe~\cite{pennington2014glove}, which is the only pretrained word embeddings used in previous depression detection tasks~\cite{Williamson:2016:DDU:2988257.2988263,bjorn2019}, BERT and ELMo are believed to be more context-aware. 
Therefore, the use of pretrained contextual sentence embeddings is investigated in the current work for their usage in depression detection.

\section{Approach}
\label{sec:approac}
In this section, we detailed our novel approach of using a multi-task setting to model depression presence and severity prediction tasks together, with the use of only text features for neat real-world applications. 

\subsection{Multi-task modeling}
Prior work on depression detection usually splits the tasks of depression presence detection (binary classification)~\cite{Yang:2016:DTB:2988257.2988269} and severity score prediction (regression with PHQ-8 score)~\cite{Scherer:2014:DBA:2663204.2663238}.
However, as discussed previously, the two characteristics are correlated, but one cannot necessarily predict the other. 
Hence, both information sources are essential in order to ascertain the patients' state.

\begin{align}
    \ell_{bce}\left(x,y\right) &= -\left[ y \cdot \log x+(1-y) \cdot \log (1-x) \right]\label{eq:bce} \\
    \ell_{hub}\left(x,y\right) &=
        \begin{cases}
        0.5 (x - y)^2, & \text{if } |x - y| < 1 \\
        |x - y| - 0.5, & \text{otherwise }
        \end{cases}\label{eq:huber}
\end{align}

We thus propose a multi-task setting to combine the classification and regression tasks.
Two outputs are constructed, one directly predicts the binary outcome of a participant being depressed, the other outputs the estimated PHQ-8-score. 
We opt to use a combination of binary cross entropy (BCE, for classification, \Cref{eq:bce}) and huber loss (for regression, \Cref{eq:huber}) in our work. 
The Huber loss can be seen as a compromise between mean average error (MAE, L1) and mean square error (MSE, L2), resulting in a robust behavior to outliers.
While in theory, $\ell_{bce} + \ell_{hub}$ is a reasonable choice for a loss function, in practice $\ell_{hub}$ dominates $\ell_{bce}$, meaning that the model is likely to focus on regression rather than the binary classification.

\begin{align}
    \ell(x_r,x_c,y_r, y_c) &= \label{eq:loss} \\
        (1-w)\ell_{bce}(\sigma(x_c),y_c)  +  w\ell_{hub}(x_r, y_r) & \notag
\end{align}

To alleviate this problem, we introduce a fixed weight factor $w$, to create the convex combination loss $\ell$ (\Cref{eq:loss}).
During training, $w$ is set to $0.1$.
Here, $x_r$ represents the regressive model output, $x_c$ represents the binary model output, $\sigma$ is the sigmoid function, $y_r$ is the PHQ-8 score, and $y_c$ is the binary ground truth.

\subsection{Pooling method}

Since labels for this task are only given per interview, meaning after a sequence of questions and answers, a pooling layer is required to remove all time-variance to a single vector representation $\mathbf{z}$ and evaluate the entire dialogue.
Pooling methods can be sub-categorized into hidden-level and output-level approaches.
Hidden-level pooling reduces an intermediate representation, e.g., the BGRU output ($\mathbf{O}$), while output-level reduces per-timestep probability predictions to one, e.g., after a soft-max layer. 
In literature, it has been observed that hidden-level pooling, specifically in closely related tasks such as sound event detection and speaker verification is superior to output-level ones~\cite{Lin2019,Kao_rare2020,x_vector,Variani-icassp14,Wang2018}.
We, therefore, focus on hidden-level pooling methods.
This paper models depression detection as a sequence of text-embeddings $\mathbf{X} = \left[\mathbf{x}_1, \ldots, \mathbf{x}_T \right]$, either on word- or sentence-level and their corresponding BGRU representations $\mathbf{O} = \left[\mathbf{o}_1, \ldots, \mathbf{o}_T \right]$, where $\mathbf{x}_t$ represents a text-embedding at time $t$.

\begin{figure}
    \centering
    \includegraphics[width=0.8\linewidth]{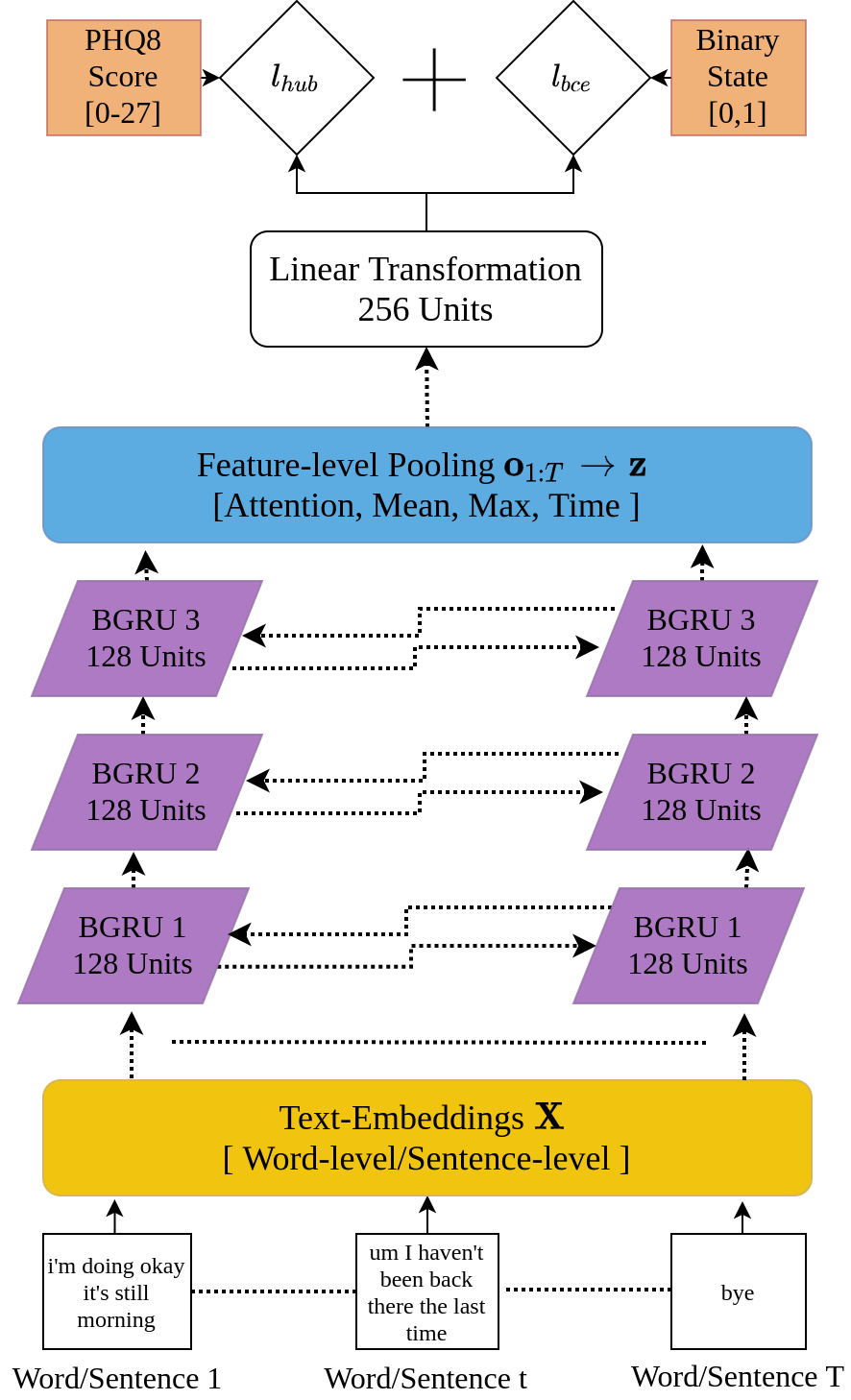}
    \caption{Proposed model architecture. The output of the last layer are two values, one for the regression ($x_r$, PHQ-8) and one for classification ($x_c$). The architecture consists of 3 layer BGRU model with 128 hidden units.}
    \label{fig:blstm_arch}
\end{figure}

\begin{table}[htpb]
    \centering
    \def\arraystretch{1.2}
    \begin{tabular}{l|r}
    \toprule
        Name & Function \\
        \hline\hline
        Time & $\mathbf{z}_{T} = \mathbf{o}_T$   \\
        Mean &  $\mathbf{z}_{mean} = \frac{1}{T} \sum^T \mathbf{o}_t$\\
        Max  & $\mathbf{z}_{max} = \max_{t}(\mathbf{o}_{1 : T})$\\
        Attention & $\mathbf{z}_{att} = \sum_t^T \mathbf{\alpha}_t \mathbf{o}_t$\\
        \bottomrule
    \end{tabular}
    \caption{Pooling functions utilized in this work.}
    \label{tab:pooling_functions}
\end{table}
Previous text-based work in~\cite{AlHanai2018, Haque2018} solely relied on the last-timestep ($\mathbf{z}_{T}$), further referred to as time-pooling, or mean-pooling ($\mathbf{z}_{mean}$) methods as the response/query representation. 
However,~\cite{Mirsamadi2017} has shown that time-pooling is only sub-optimal since the network belief changes over time. 
In this work, we investigate the usage of four different pooling functions, seen in \Cref{tab:pooling_functions}.
All approaches with the exception of attention are parameter-free, which is potentially helpful in sparse data scenarios.
The individual attention weights ($\alpha_t$) are estimated given the concatenated forward and backward hidden states from the BGRU model at time $t$:
\[
    \alpha_t = \frac{e^{\mathbf{v}\mathbf{o}_t}}{\sum_j e^{\mathbf{v}\mathbf{o}_j}}\\
\]
Here $\mathbf{v}$ is the trainable time-independent attention weight vector.
In addition to the novel multi-task approach and attention pooling method stated above, our proposed architecture in this work is a commonly used bidirectional gated recurrent unit (BGRU)  neural network structure (see \Cref{fig:blstm_arch}). 
After each BGRU layer, we apply a recurrent dropout with a probability of 20 \%.
In our initial experiments, we also investigated long short term memory (LSTM) networks. 
Even though the best performance achieved is on par with BGRU models, average performance is significantly worse, likely due to the additional number of parameters in a BLSTM model.
The source code is publicly available.\footnote{https://gitlab.com/Richy/text-based-depression-detection}

\subsection{Text embeddings}
\label{ssec:embeddings}

In previous depression detection research, context-free word embeddings are usually used, either trained from scratch with Word2Vec~\cite{le2014distributed} or a simple pretrained word embedding GloVe~\cite{pennington2014glove}.
The most commonly used Word2Vec/Doc2Vec models are usually trained on a shallow, two-layer deep neural network architecture. 
While Word2Vec aims to capture the context of a specific sentence, it only considers the surrounding words as its training input; therefore, it does not capture the intrinsic meaning of a sentence.

\paragraph*{Pretraining}
Since depression data is hard to come by, using a model pretrained on large text corpora could help alleviate data sparsity problems.
In our work, we mainly focus on analyzing pretrained word-embeddings against their from-scratch counterparts.

\paragraph*{Word and Sentence-embeddings}
Text-embeddings can be extracted on multiple, abstract levels such as character, sub-word, word, sentence, and paragraph (Doc2Vec).
Traditional methods such as Word2Vec are generally extracted on word-level, meaning that each word is represented by a $D$ dimensional feature vector.
Here, our focus is to compare word-level embeddings to sentence-level ones.
Our assumption is that word-level embeddings are unfit to model depression detection in a sequential fashion since each word contains little information about the entire context of a session.
Moreover, the sequence length of the feature $\mathbf{X}$ on word-level is much larger than on sentence-level.
We, therefore, propose to compare two variations of our utilized word-level features.
Sentence-level Word2Vec/fastText embeddings (Word2Vec (S)/fastText (S)) from a sentence $j$ are extracted from their word-level representations $\mathbf{w}_t$ as:
\[
    \mathbf{x}_{j} = \frac{1}{T_j} \sum_{t=1}^{T_j} \mathbf{w}_t 
\]
where $T_j$ is the number of words within that sentence.
In contrast, modern text-embeddings such as ELMo and BERT are context-sensitive, meaning they produce at least sentence-level representations.
Recently, models like BERT~\cite{devlin2018bert} and ELMo~\cite{Peters:2018} made use of the self-attention mechanism and LSTMs to build context-sensitive sentence representations. 
ELMo and BERT models pretrained on large corpus such as Wikipedia are publicly available, therefore, can be effectively used as high-level feature extractors.
Both models generate embeddings for a word based on the context it appears in, thus produces slight variations for each word occurrence. 
Subsequently, they require to be fed an entire sentence before generating an embedding.
Therefore, they fundamentally differ from traditional Word2Vec embeddings, which can create a sentence embedding as an average of all word embeddings. 
Four different text embeddings are thus experimented with: Word2Vec, fastText, ELMo, and BERT:

\begin{figure*}
    \centering
    \includegraphics[width=\linewidth]{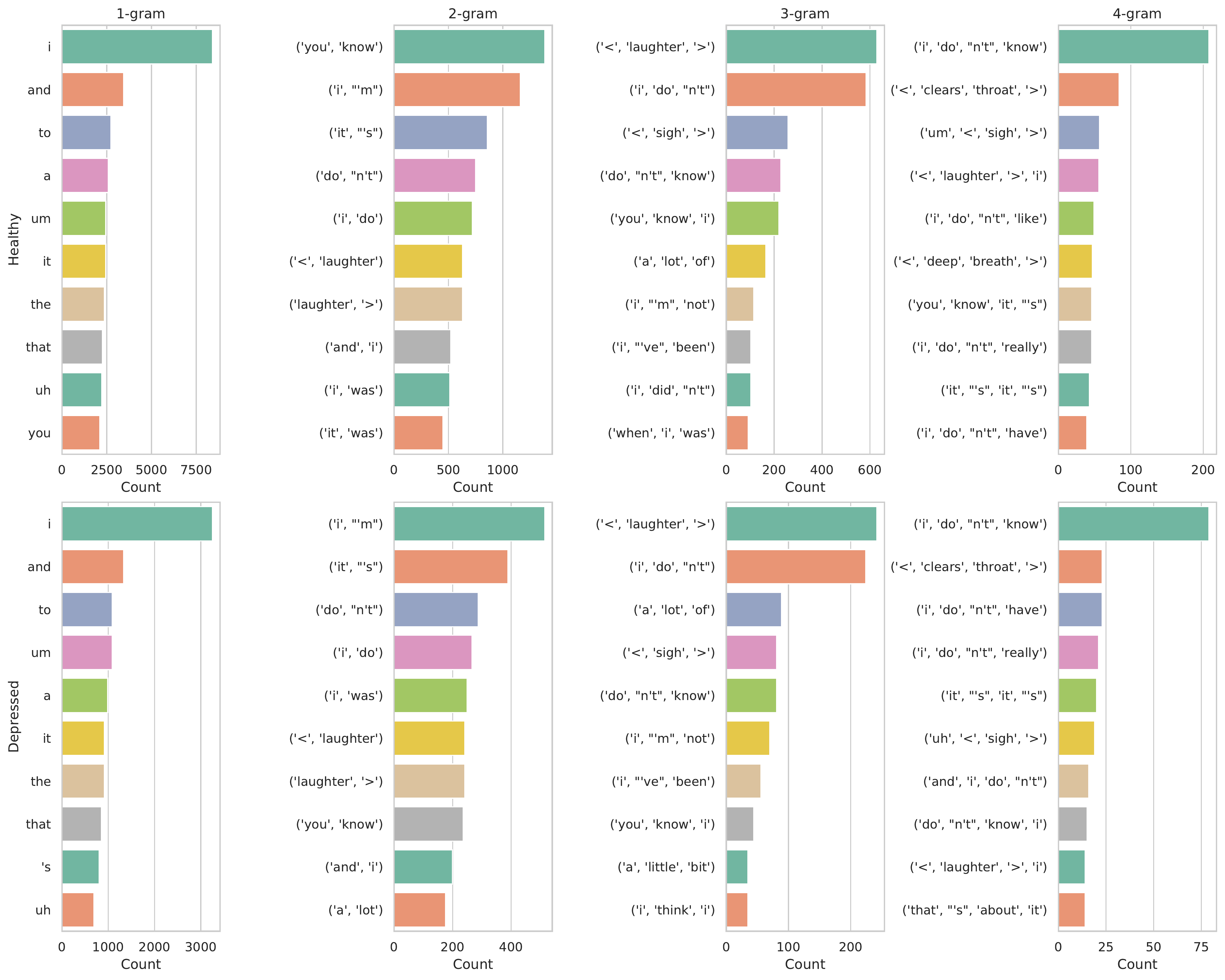}
    \caption{Top-10 Training data n-grams. Top row: Healthy patients; Bottom row: Depressed patients. Best viewed in color.}
    \label{fig:train_ngram_dist}
\end{figure*}

\textbf{Word2Vec} A Word2vec~\cite{mikolov2013efficient} model was trained with the hyper-parameters dimension $D=300$, minimum word count 5, number of iterations 5, using hierarchical softmax as well as skip-ngram.
For training from scratch (\xmark), the entire DAIC-WOZ~\cite{Ringeval:2017:ARD:3133944.3133953} text corpus was utilized.
For experiment utilizing pretrained (\cmark) embeddings, an up-to-date Wikipedia dump (16 GB) was utilized as the dataset.
The entire process is implemented using the gensim library~\cite{rehurek_lrec}.

\textbf{fastText} Another popular text-embedding is fastText, which different from Word2Vec is sub-word based (e.g., consecutive sequences of characters).
A fastText~\cite{joulin2016bag,bojanowski2017enriching} model was trained with the hyper-parameters dimension $D=300$, minimum word count 5, number of iterations 8, using hierarchical softmax as well as skip-ngram.
Like Word2Vec, training from scratch used the entire DAIC-WOZ dataset, while fastText pretraining was done on Wikipedia. 

\textbf{ELMo} Pretrained ELMo embeddings are utilized in this work. 
The high-level ELMo representation is based on low-level character-level inputs; thus, in theory, ELMo should be able to handle out-of-vocabulary (OOV) words better than other approaches, specifically non-verbal information such as ``laughter''.
ELMo uses a three-layer bidirectional structure with $1024$ nodes in each layer. 
The model was pretrained on the 1 billion word benchmark dataset. 
We used the average of all three layer embeddings as our sentence representation. 

\textbf{BERT} Currently, multiple BERT versions, each of different size are commonly available~\cite{devlin2018bert}. 
Here, we utilize the basic 12 layers, 768 hidden size representation, Bert-Base model as our standard BERT representation model.
An embedding can be extracted from each layer. 
The official Bert-Base model has been pretrained on the Wikipedia dataset.
The penultimate BERT model layer was used to extract a single $768$ dimensional sentence embedding.
A maximum sequence length of $125$ was set in order to reduce memory consumption.

\begin{table}[htpb]
    \centering
    \begin{tabular}{l|rr}
    \toprule
        Level & Embedding & Dimension \\
        \hline\hline
        \multirow{2}{*}{Word} & Word2Vec (W) & 300  \\
        & fastText (W) & 300  \\
        \hline
        \multirow{4}{*}{Sentence} & Word2Vec (S) & 300  \\
        & fastText (S) & 300  \\
        & BERT & 768    \\
        & ELMo & 1024 \\
    \bottomrule
    \end{tabular}
    \caption{Text-embedding dimensions utilized in this work.}
    \label{tab:text_features}
\end{table}

All utilized features with regards to their input dimension $D$ can be seen in \Cref{tab:text_features}.

\section{Experiments}
\label{sec:experiments}
\subsection{Experimental Setting}
\noindent\textbf{Dataset}
Data was acquired from the publicly available WOZ-DAIC~\cite{Gratch_thedistress,DeVault:2014:SKV:2615731.2617415} corpus, which encompasses three major media: video, audio, and manually transcribed text data. 
All modalities are manually labeled, meaning that the available data is of higher quality than, e.g., automatically transcribed data by a machine.
This also means that our work can be seen as a possibly optimal performance towards automatic depression detection since real-world approaches will require automatic labeling, thus likely having erroneous transcriptions/labels.
Prior work on this dataset generally focuses on utilizing modality fusion methods~\cite{AlHanai2018,Haque2018,Williamson:2016:DDU:2988257.2988263}, and it is suggested that a critical factor for depression detection is the addition of semantic text information.
An evaluation subset was also published, yet labels for the evaluation are not available. Therefore all experiments were validated on the development subset. 
We first analyze the data regarding any connection between depressed and healthy patients in terms of the provided transcripts.
The n-gram counts for the training subset can be seen in \Cref{fig:train_ngram_dist}.
Multiple observations from the n-grams can be made: 1) The difference in content words between healthy and depressed patients is little. 2) Three- and four-grams both contain a high count of behavioral states, e.g., laughter, sign, clears throat. 3) Most n-grams do not contain meaningful information, e.g., ``I don't know, I don't have, I do''.

\begin{figure}
    \centering
    \includegraphics[width=\linewidth]{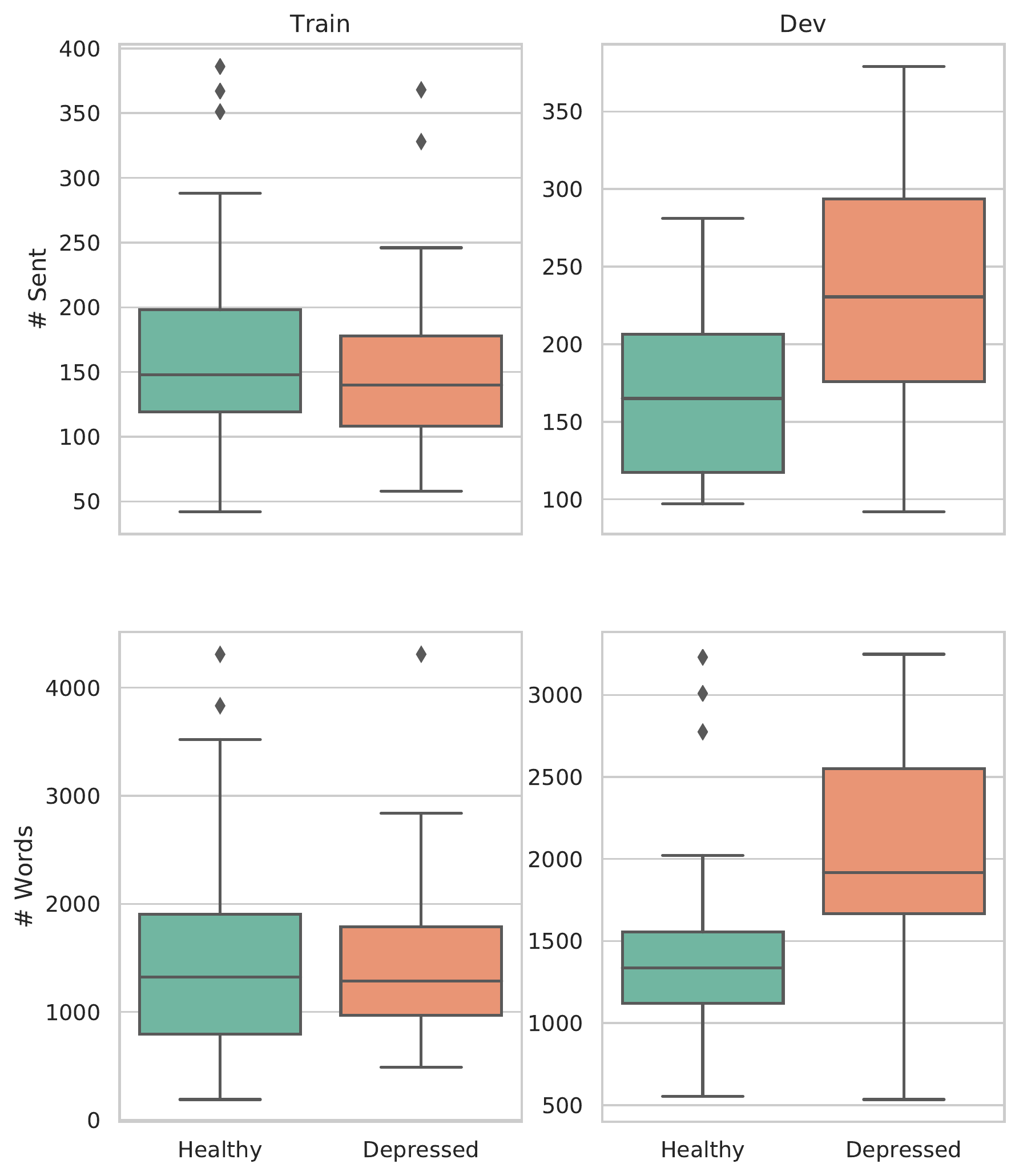}
    \caption{Sentence and word length distribution between the training (left) and development (right) dataset for depressed and healthy patients.} 
    \label{fig:train_dev_word_sent_count}
\end{figure}

\hfill \break
\noindent\textbf{Data preprocessing} Our approach used a sequential modeling method, which processes the patients' response in succession without any regard to interactions within the questioner.
The raw text was firstly preprocessed, where tailing blanks were removed, and every letter was set to be lowercase.
The preprocessed training set contains an overall $16895$, the development set $6674$ written sentences.
The training set contains $107$, the development set $35$, patients.
The average number of sentences in each subset are $157$ and $190$ for training and development subset respectively (see \Cref{fig:train_dev_word_sent_count}).

Our sequential modeling approach treats an entire paragraph spoken by a patient as a single sample. Thus the training data only contains $107$ samples.
Meta information such as \textless laughter\textgreater or \textless sigh\textgreater are possibly helpful to the model, thus were kept. 
The training and development dataset distribution regarding sentences and words can be seen in \Cref{fig:train_dev_word_sent_count}.
It can be observed that the training set distributions for the number of words as well as the number of sentences are similar for depressed as well as healthy patients.
However, in the development set, the depressed patient distribution largely shifts towards much longer sentences and higher words counts.
In order to ascertain reproducible and meaningful results, all experiments utilize a stratified 5-Fold cross-validation scheme~\cite{sechidis2011stratification}.
For each experiment, we first divide the $107$ available training-data samples (patients) into 86 (80\%) samples used for model training and cross-validate (cv) on the rest 21 samples (20\%).
Due to stratification, we assure that each folds' binary state distribution is roughly similar.
The seed generating each fold is fixed for all experiments, meaning that the main difference between consecutive runs is the model parameter initialization.

\hfill \break
\noindent\textbf{Neural network training} Training was done by running Adam~\cite{KingmaB14} optimization for at most $200$ epochs.
The initial learning rate was set to be $0.004$, which was reduced by a factor of $10$ if the cross-validation loss did not improve for at most $3$ epochs. 
Early stopping was utilized, whereas training stops if no loss improvement has been seen for at most $10$ epochs. 
The model producing the lowest loss on the held-out cross-validation set was chosen for evaluation on the development (dev) set. 

Regarding data handling, padding was applied on batch-level by padding zeros towards the longest utterance within a batch.
In order to avoid any influence stemming from the padded sequences, all pooling functions are implemented using a mask, which neglects padded elements during computation of $\mathbf{z}$.
A batch-size of 32 was utilized in this work.
Due to the inherent imbalance between depressed and healthy samples ($\approx 2:1$ ratio), random oversampling over the minority class (depressed samples) was utilized.
Further prevention of over-fitting was done by assuring a balanced sample distribution (1:1) within each batch.
Recurrent weights were initialized by the uniform Xavier~\cite{Glorot10understandingthe} method, where samples were drawn from $[-\beta,\beta]$, where $\beta=\sqrt{\frac{6}{\mathbf{W}_{\text{in}} + \mathbf{W}_{\text{out}}}}$ and biases were set to zero.
Initial weights regarding the attention parameter $\mathbf{v}$ were drawn from a normal distribution $\mathcal{N}(0,0.05)$.
Pseudo-random generator seeds for each respective experiment were fixed regarding weight initialization and data sampling.
Deep learning models generally require a large data corpus in order to work. 
Since the available amount of data can be considered insufficient, the parameter initialization, as well as the choice of hyper-parameters, is vital.
In order to circumvent this problem~\cite{AlHanai2018} proposed to grid-search for every possible hyper-parameter aiming to ascertain a proper configuration.
In this work, a hyper-parameter search was also conducted; thus the outlined parameters can be considered optimal in our setting.

\hfill \break
\noindent\textbf{Evaluation metric}
For classification, accuracy (Acc), macro precision (pre), and recall (rec) scores are used to calculate the F1-score.
In terms of regression, the mean average error ($|x-y|$) and root mean square error ($\sqrt{\left(x-y\right)^2}$) are used between the model prediction $x$ and the ground truth PHQ-8 score $y$.
Our \textit{primary} metrics for most experiments are F1, Acc and MAE. 
Our \textit{secondary} metrics, additionally use precision, recall and RMSE.
Note that previous work obtained the classification performance by thresholding the PHQ-8 regression result with a value of $10$, while our approach decouples classification and regression performance.
For all experiments, if not further specified, we report average scores after 100 runs on each fold.

\subsection{Comparison Methods}
\noindent\textbf{Text settings} Apart from sequence modeling, two other different settings~\cite{AlHanai2018} are widely used:

\textit{Context-free} modeling uses each response of the participant as an independent sample, without information about the question, nor the time it was asked. 

\textit{Context-dependent} modeling requires the use of question-answer pairs, where each sample consists of a question asked and its corresponding answer~\cite{gong2017topic}.

\section{Results}
\label{sec:results}

In this section, we provide our results and aim to interpret the data.
Our baseline pooling method for the following experiments is mean-pooling.

\subsection{Pretrained vs. from scratch on word-embeddings}

Our first experiment aims to investigate the use of pretrained word-level embeddings (here \cmark) for depression detection.
Word2Vec and fastText are chosen as our representative embeddings (see \Cref{tab:text_features}).
Embeddings from scratch (here \xmark) refer to training a Word2Vec/fastText model on the provided DAIC dataset.
In order to simplify the results, we only utilize macro-F1, MAE, and accuracy as our main metrics.
This experiment uses mean-pooling and each reported metric $\rho$ having value $r$ of fold $f$ and run $k$ is calculated as:
\[
    \mu_{f}(\rho) = \frac{1}{K} \sum_{k=1}^{K=100} r_f^k(\rho)
\]
The Word2Vec results can be seen in \Cref{tab:word2vec_scratch_vs_pre} and fastText results in \Cref{tab:fastText_scratch_comp}.

\begin{table}
\centering
\begin{tabular}{ll|cc|cc|cc}
\toprule
       &     &  \multicolumn{2}{c}{F1} &  \multicolumn{2}{c}{MAE} & \multicolumn{2}{c}{Acc} \\
   Fold & Data & \xmark & \cmark & \xmark & \cmark  &  \xmark & \cmark   \\
\midrule
\multirow{2}{*}{1} & cv & 0.36 & 0.35 & 4.19 & 4.28 &  0.55 &    0.52 \\
                   & dev & 0.42 & 0.40 & 5.65 &  5.68 & 0.56 &    0.55 \\
\multirow{2}{*}{2} & cv & 0.34 & 0.46 & 4.65 & 4.48 &    0.52 &  0.58 \\
                   & dev & 0.35 & 0.49 & 5.63 & 5.28 &     0.52 & 0.60 \\
\multirow{2}{*}{3} & cv & 0.33 & 0.38 & 3.82 & 3.80 & 0.50 &    0.56 \\
                   & dev & 0.33 & 0.40 & 5.66 & 5.59 &    0.50 & 0.56 \\
\multirow{2}{*}{4} & cv & 0.33 &  0.50 & 5.31 & 4.23 &  0.53 &    0.66 \\
                   & dev & 0.33 & 0.52 & 5.60 & 5.15 & 0.52 &  0.62 \\
\multirow{2}{*}{5} & cv & 0.35 & 0.44 & 6.32 & 5.76 & 0.55 &    0.64 \\
                   & dev & 0.35 & 0.48 & 5.69 &  5.28 & 0.53 &   0.61 \\
\hline
\multirow{2}{*}{Avg} & cv & 0.34  & \textbf{0.42} & 4.86 & \textbf{4.50} & 0.53 & \textbf{0.59} \\
                &  dev & 0.36 & \textbf{0.45} & 5.67 & \textbf{5.39} & 0.53 & \textbf{0.58} \\
\bottomrule
\end{tabular}
\caption{Comparison between Word2Vec embeddings trained from scratch (\xmark) on the provided DAIC-WOZ data and pretrained (\cmark) on the Wikipedia dataset.
Values represent the average performance ($\mu_f(\rho)$) for each fold.
Best in bold.
}
\label{tab:word2vec_scratch_vs_pre}
\end{table}

The results show that specifically, MAE scores between cv and dev datasets largely differ. 
This is likely due to the great difference in sentence length and word counts between cv (train) and development subsets (see \Cref{fig:train_dev_word_sent_count}).
Similar discrepancies between cv and development sets have also been reported~\cite{gong2017topic}.
Further, both utilized front-ends (Word2Vec/fastText), improve across all utilized metrics when using pretrained embeddings.
Specifically, Word2Vec significantly enhances its performance on the development set (F1 0.36 $\rightarrow$ 0.45, MAE 5.67 $\rightarrow$ 5.39) when pretraining is utilized.
fastText also improves across the F1 and Acc metrics, yet its performance gains are smaller than Word2Vec ones (F1 0.35 $\rightarrow$ 0.41).
Interestingly, fastTexts' MAE results on the development set slightly drops when pretraining is utilized.
However, since its classification performance, especially regarding accuracy, is greatly enhanced, we believe that fastText could potentially be a better feature for classification than regression.

\begin{table}
\centering
\begin{tabular}{ll|cc|cc|cc}
\toprule
       &     &  \multicolumn{2}{c}{F1} &  \multicolumn{2}{c}{MAE} & \multicolumn{2}{c}{Acc} \\
   Fold & Data & \xmark & \cmark & \xmark & \cmark  &  \xmark & \cmark   \\
\midrule
\multirow{2}{*}{1} & cv & 0.36 & 0.35  & 4.17 & 4.22 &     0.56 & 0.51 \\
                   & dev & 0.38 &  0.40 & 5.65 & 5.66 & 0.55 & 0.54 \\
\multirow{2}{*}{2} & cv & 0.36 & 0.42 &  4.64 & 4.58 &     0.53 & 0.58 \\
                   & dev& 0.37 & 0.44 & 5.63 & 5.53 &     0.53 & 0.58 \\
\multirow{2}{*}{3} & cv & 0.37 & 0.45 &  3.82 & 3.80 &     0.56 & 0.61 \\
                  & dev & 0.36 & 0.41 & 5.63 & 5.66  &     0.54 & 0.57 \\
\multirow{2}{*}{4} & cv & 0.34 & 0.40 & 5.24 &  4.86 &     0.53 & 0.57 \\
                  & dev & 0.34 & 0.41 & 5.68 & 5.53  &  0.52 & 0.54 \\
\multirow{2}{*}{5} & cv & 0.34 & 0.38 & 6.29 & 6.02 &  0.52 & 0.60 \\
                  & dev & 0.34 & 0.40 & 5.68 & 5.56 &  0.51 & 0.57 \\
                   \hline
\multirow{2}{*}{Avg} & cv & 0.35  & \textbf{0.39} & 4.83 & \textbf{4.69} & 0.53 & \textbf{0.57} \\
                &  dev & 0.35 & \textbf{0.41} & \textbf{5.58} & 5.65 & 0.53 & \textbf{0.65} \\
\bottomrule
\end{tabular}
\caption{Comparison between fastText embeddings trained from scratch (\xmark) on the provided DAIC-WOZ data and pretrained (\cmark) on the Wikipedia dataset.
Values represent the average performance ($\mu_f(\rho)$) for each fold.
Best in bold.
}
\label{tab:fastText_scratch_comp}
\end{table}

Since conclusions drawn from these experiments consistently demonstrate performance gain benefiting from pretrained text embeddings, the following experiments by default use pretrained embeddings.

\subsection{Word- vs. sentence-level embeddings}

Regarding different text embeddings, another critical question is, do sentence-level features enhance performance compared to word-level ones?
Our BGRU model is capable of sequence modeling, while lacking at word-level processing, since each word contributes little information to the entire session's entire context.
A popular approach to enhance partial information modeling is chunking~\cite{zhai_chunking}.
In our case, we see chunking as the word-level averaging within a sentence, as previously introduced in \Cref{ssec:embeddings}.

\begin{table}
\centering
\begin{tabular}{ll|cc|cc|cc}
\toprule
       &     &  \multicolumn{2}{c}{F1} &  \multicolumn{2}{c}{MAE} & \multicolumn{2}{c}{Acc} \\
   Fold & Data & W & S &     W & S  &  W & S   \\
\midrule
\multirow{2}{*}{1} & cv & 0.35 & 0.33  & 4.28 & 4.35 &     0.52 & 0.49 \\
                   & dev & 0.40 &  0.36 & 5.68 & 5.71 & 0.51 & 0.55 \\
\multirow{2}{*}{2} & cv & 0.46 & 0.62 &  4.48 & 3.95 &     0.58 & 0.51 \\
                   & dev & 0.49 & 0.59 & 5.28 & 4.85 &     0.60 & 0.67 \\
\multirow{2}{*}{3} & cv & 0.38 & 0.43 &  3.80 & 3.72 &     0.56 & 0.53 \\
                   & dev & 0.40 & 0.49 & 5.59 & 5.29  &     0.56 & 0.58 \\
\multirow{2}{*}{4} & cv & 0.50 & 0.64 & 4.23 &  3.36 &     0.66 & 0.79 \\
                   & dev & 0.52 & 0.64 & 5.15 & 4.71 & 0.62 & 0.70 \\
\multirow{2}{*}{5} & cv & 0.44 & 0.56 & 5.76 & 5.19 &  0.64 & 0.73 \\
                   & dev & 0.48 & 0.58 & 5.28 & 4.90 &  0.61 & 0.67 \\
                   \hline
\multirow{2}{*}{Avg} & cv & 0.42  & \textbf{0.51} & 4.50 & \textbf{4.11} & 0.59 & \textbf{0.64} \\
                &  dev & 0.45 & \textbf{0.53} & 5.39 & \textbf{5.09} & 0.58 & \textbf{0.63} \\
\bottomrule
\end{tabular}
\caption{Comparison between Word2Vec pretrained embeddings on word (W) and sentence (S) level using mean-pooling. 
Values represent the average performance ($\mu_f(\rho)$) for each fold.
Best in bold.}
\label{tab:word2vec_pretrain_comparison}
\end{table}

Our experimental results comparing word-level (W) and sentence-level (S) embeddings can be seen in \Cref{tab:word2vec_pretrain_comparison} and \Cref{tab:fastText_comparison}.
For both front-end features, considerable improvements are observed when using sentence-level features instead of word-level ones.
Word2Vec performance regarding F1 (0.45 $\rightarrow$ 0.53), accuracy (0.58 $\rightarrow$ 0.63) and MAE (4.50 $\rightarrow$ 4.11) all significantly benefit from sentence-level features.
Like our initial pretraining experiments using fastText, its performance also improves, yet it is less notable than Word2Vec.
We, therefore, continue to compare sentence-level features towards their usage in depression detection.

\begin{table}
    \centering
\begin{tabular}{ll|cc|cc|cc}
\toprule
       &     &  \multicolumn{2}{c}{F1} &  \multicolumn{2}{c}{MAE} & \multicolumn{2}{c}{Acc} \\
   Fold & Data & W & S &     W & S  &  W & S   \\
\midrule
\multirow{2}{*}{1} & cv & 0.35 & 0.33  & 4.22 & 4.27 &     0.51 & 0.51 \\
                   & dev & 0.40 &  0.36 & 5.68 & 5.71 & 0.51 & 0.55 \\
\multirow{2}{*}{2} & cv & 0.42 & 0.40 &  4.58 & 4.57 &     0.58 & 0.56 \\
                   & dev & 0.44 & 0.39 & 5.53 & 5.52 &     0.58 & 0.56 \\
\multirow{2}{*}{3} & cv & 0.45 & 0.45 &  3.80 & 3.74 &     0.61 & 0.61 \\
                  & dev & 0.41 & 0.45 & 5.66 & 5.48  &     0.57 & 0.59 \\
\multirow{2}{*}{4} & cv & 0.40 & 0.41 & 4.86 &  4.60 &     0.57 & 0.64 \\
                  & dev & 0.41 & 0.47 & 5.53 & 5.31 &      0.54 & 0.62 \\
\multirow{2}{*}{5} & cv & 0.38 & 0.41 & 6.02 & 5.58 &  0.60 & 0.64 \\
                  & dev & 0.40 & 0.50 & 5.56 & 5.14 &  0.57 & 0.63 \\
                   \hline
\multirow{2}{*}{Avg} & cv & \textbf{0.39}  & \textbf{0.39} & 4.69 & \textbf{4.57} & 0.57 & \textbf{0.58} \\
                &  dev & 0.41 & \textbf{0.42} & 5.58 & \textbf{5.43} & 0.55 & \textbf{0.58} \\
\bottomrule
\end{tabular}
\caption{Comparison between fastText pretrained embeddings on word (W) and sentence (S) level using mean-pooling. Each fold is run 100 times and the mean values for all experiments are reported. }
    \label{tab:fastText_comparison}
\end{table}

\subsection{Comparing pooling methods}

In this experiment we aim to ascertain the usage of the four proposed pooling functions, namely Attention, Max, Mean, and Time pooling.
We report our findings (metric $\rho$) as a global average ($\mu$) across all runs, independent of folds:
\[
    \mu(\rho) = \frac{1}{FK} \sum_{f=1}^{F=5}  \sum_{k=1}^{K=100} r_f^k(\rho)
\]
The results are displayed in \Cref{tab:pooling_comp}.
Time-pooling can be seen to be the least favorable pooling method, regardless of which text embedding is utilized.
The other three pooling methods' performance are slightly different depending on the front-end embeddings.  
Across all utilized pooling functions, ELMo is seen to provide consistently well performing results regarding F1 and MAE.
Utilizing ELMo with attention, mean or max pooling results in an average MAE of $\approx 5.06$.
Since ELMo is pretrained on a larger dataset (1 Billion word) compared to the other embeddings, we believe that this is the main reason for the performance gains on the development set.
BERT embeddings, while outperforming ELMo ones on the cv dataset, degrade on the development set.
Interestingly, Word2Vec embeddings with mean-pooling are observed to produce the lowest MAE error in the held-out cv set.
We conclude from the set of experiments, that the pooling functions are essential, since by choosing the naive time-pooling strategy, performance significantly drops.
Mean, attention and max pooling all seem to be a viable approach for depression detection.

\begin{table}
\centering
\begin{tabular}{lllrrr}
\toprule
Data & Pooling & Embedding &  MAE    &  F1    & Acc         \\
\midrule
\multirow{16}{*}{cv} & \multirow{4}{*}{Att} & BERT & 4.16 & \textbf{0.53} &     \textbf{0.69} \\
    &      & ELMo & 4.31 & 0.50 &     0.66 \\
    &      & fastText (S) & 4.20 & 0.50 &     0.67 \\
    &      & Word2Vec (S) & 4.14 & \textbf{0.53} &     0.66 \\
    \cline{2-6}
    & \multirow{4}{*}{Max} & BERT & 4.21 & 0.50 &     0.68 \\
    &      & ELMo & 4.24 & 0.52 &     0.64 \\
    &      & fastText (S) & 4.49 & 0.45 &     0.63 \\
    &      & Word2Vec (S) & 4.15 & 0.51 &     0.62 \\
    \cline{2-6}
    & \multirow{4}{*}{Mean} & BERT & 4.21 & 0.51 &     0.68 \\
    &      & ELMo & 4.28 & 0.51 &     0.67 \\
    &      & fastText (S) & 4.57 & 0.40 &     0.59 \\
    &      & Word2Vec (S) & \textbf{4.11} & 0.52 &     0.65 \\
    \cline{2-6}
    & \multirow{4}{*}{Time} & BERT & 4.66 & 0.43 &     0.65 \\
    &      & ELMo & 4.75 & 0.41 &     0.64 \\
    &      & fastText (S) & 4.62 & 0.42 &     0.66 \\
    &      & Word2Vec (S) & 4.62 & 0.41 &     0.66 \\
    \hline
\multirow{16}{*}{dev} &  \multirow{4}{*}{Att} & BERT & 5.21 & 0.51 &     0.64 \\
    &      & ELMo & 5.08 & 0.55 &     0.66 \\
    &      & fastText (S) & 5.26 & 0.51 &     0.63 \\
    &      & Word2Vec (S) & 5.13 & 0.53 &     0.64 \\
    \cline{2-6}
    & \multirow{4}{*}{Max} & BERT & 5.14 & 0.51 &     0.65 \\
    &      & ELMo & \textbf{5.03} & \textbf{0.59} & \textbf{0.68} \\
    &      & fastText (S) & 5.39 & 0.48 &     0.62 \\
    &      & Word2Vec (S) & 5.05 & 0.57 &     0.66 \\
    \cline{2-6}
    & \multirow{4}{*}{Mean} & BERT & 5.10 & 0.52 &     0.65 \\
    &      & ELMo & 5.07 & 0.54 &     0.66 \\
    &      & fastText (S) & 5.43 & 0.43 &     0.58 \\
    &      & Word2Vec (S) & 5.09 & 0.53 &     0.63 \\
    \cline{2-6}
    & \multirow{4}{*}{Time} & BERT & 6.76 & 0.34 &     0.47 \\
    &      & ELMo & 6.80 & 0.33 &     0.46 \\
    &      & fastText (S) & 6.81 & 0.33 &     0.45 \\
    &      & Word2Vec (S) & 6.80 & 0.32 &     0.43 \\
\bottomrule
\end{tabular}
\caption{Comparison between Time, Mean, Attention and Max pooling methods regarding four utilizied features. The global average values $\mu(\rho)$ for each metric $\rho$ are displayed.}
\label{tab:pooling_comp}
\end{table}

\begin{figure}
    \centering
    \includegraphics[width=\linewidth]{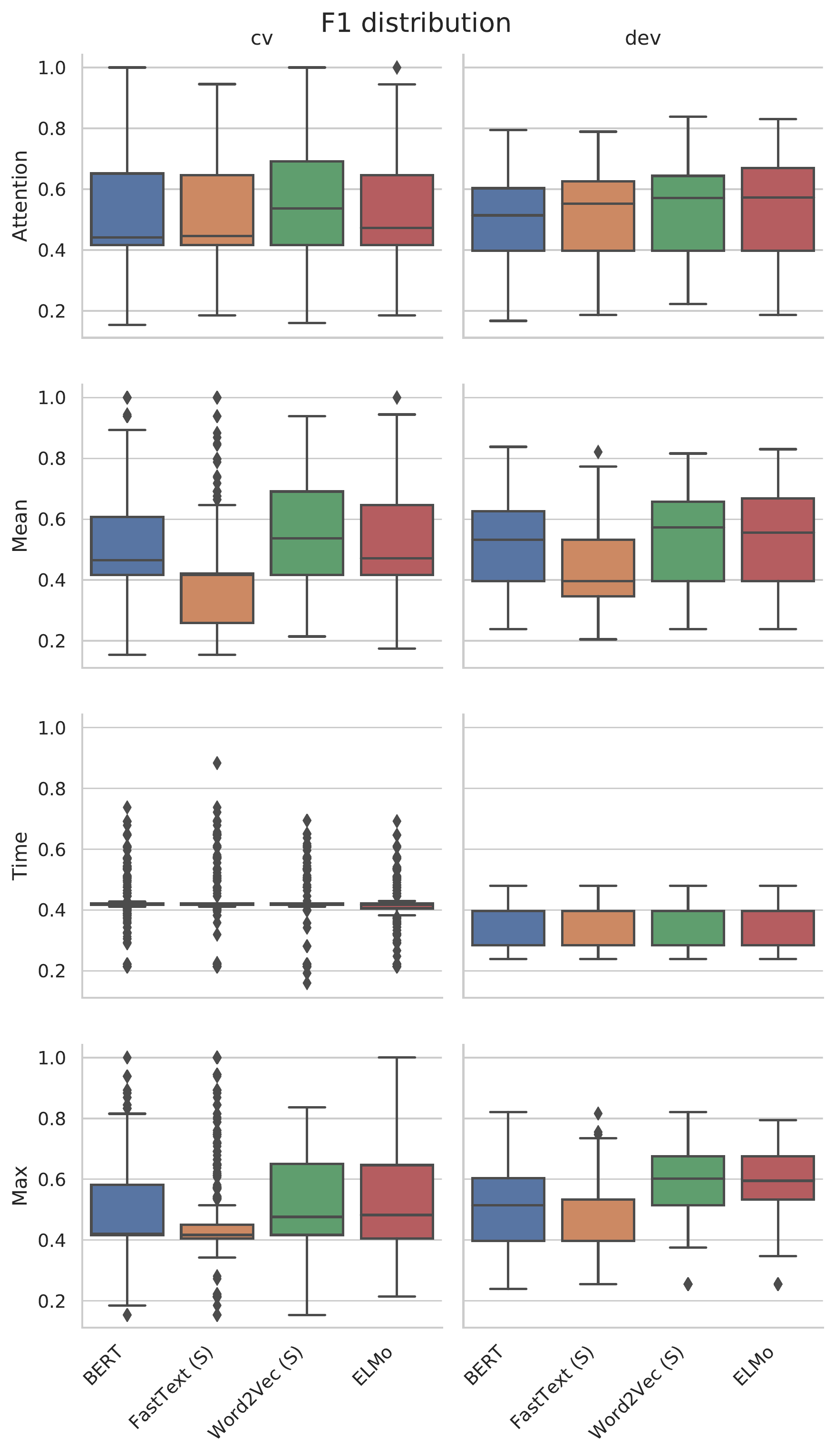}
    \caption{F1 result distribution for each single run as a boxplot for cv and development sets regarding to pooling function and text-embedding. Higher is better. Best viewed in color.}
    \label{fig:f1_distribution}
\end{figure}

\begin{figure}
    \centering
    \includegraphics[width=\linewidth]{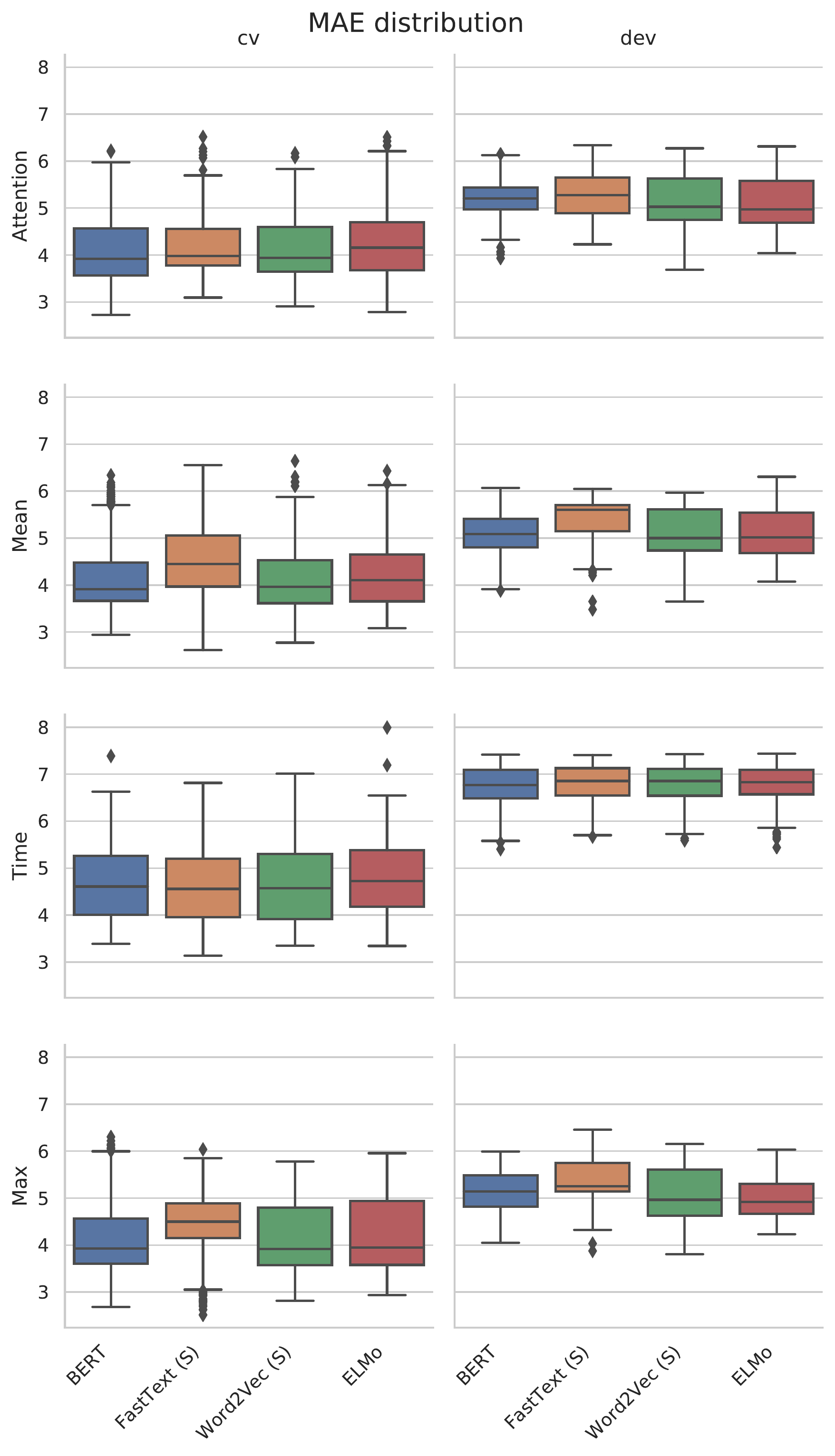}
    \caption{MAE result distribution for each single run as a boxplot for cv and development sets regarding to pooling function and text-embedding. Lower is better. Best viewed in color.}
    \label{fig:mae_distribution}
\end{figure}

\paragraph*{Result likelihood} A more detailed description of the F1 results and their significance can be seen in \Cref{fig:f1_distribution}, where each run is plotted.
Though the F1 scores distribute differently on the cv and dev sets, our model did not experience performance downfall on the dev set. 
Mainly, in terms of most stable performance, attention can be seen to produce the least amount of outliers (dots) between the cv and development sets, while having acceptable performance.
FastText is seen to produce the most outliers using mean, time, and max pooling.
The most consistent results, with the smallest variance, stem from time-pooling, even though those results are also the worst.
Different from the observed F1 scores, the MAE distribution in \Cref{fig:mae_distribution} contains overall less extreme outliers. However according to MAE results, our models are more prone to overfitting as the performance on the dev set sharply degenerated. 
Attention and max pooling methods are more stable than mean and time pooling functions. 
Consistently, time-pooling is the least favorable method investigated, regardless of the text embeddings.

\subsubsection{Best average experiment}

Lastly, we also report our best average performance ($\mu_{best}$) regarding the F1 score in \Cref{tab:best_avg_perf} calculated as:
\[
    \mu_{best} = \max_{k = 1, \ldots, 100} \left(\frac{1}{F} \sum_{f=1}^{F=5} r_f^k(F1)\right)\\
\]
Here, the best pooling method generating the result is also displayed.
As can be seen, for all features except fastText, mean-pooling indeed provides competitive performance on all data.
The most stable feature is ELMo, which offers excellent performance on the held-out cv (F1 0.73, MAE 4.26) as well as development sets (F1 0.73, MAE 4.71).

\begin{table*}[htpb]
\footnotesize
\centering
\begin{tabular}{llll|rrrr|rr}
\toprule
& & & & \multicolumn{4}{c|}{Classification} & \multicolumn{2}{c}{PHQ-8-Regression} \\
Model & Embedding &  Setting &       Pooling &   Pre &   Rec &    F1 & Acc &  MAE &  RMSE \\
\hline\hline
C-CNN \cite{Haque2018} & Word2Vec & Sequence & Time & - &  - & - & - & 6.16 & - \\
C-CNN \cite{Haque2018} & Doc2Vec & Sequence & Time &  - &  - &  - & - & 5.81 & - \\
Gauss-Staircase \cite{Williamson:2016:DDU:2988257.2988263} &  GloVe (Fusion) & Context-Dep & - & - & - & \textbf{0.84} & - & \textbf{3.34} & \textbf{4.46}  \\
Gauss-Staircase \cite{Williamson:2016:DDU:2988257.2988263} & GloVe & Context-Dep & - & - & - & 0.76 & - & - & -  \\
BLSTM \cite{AlHanai2018} & Doc2Vec & Context-Free & Time & 0.71 & 0.5 & 0.59 & - & 7.02 & 9.43 \\
BLSTM \cite{AlHanai2018} & Doc2Vec & Sequence & Time & 0.57  & 0.8 & 0.67 & - & 5.18 & 6.38  \\
GRU \cite{bjorn2019} & GloVe & Sequence & HAtt & - & 0.60 & 0.60 & - & - & - \\
\hline
BGRU & BERT &  Sequence &   Mean &       0.85 &    0.83 & \textbf{0.84} &      0.86 & 3.91 &  5.09 \\
BGRU & ELMo &  Sequence &   Att &       0.87 &    0.81 & 0.83 &      0.86 & 4.73 &  5.62 \\
BGRU & fastText (S) &   Sequence & Mean &       0.82 &    0.85 & 0.82 &      0.83 & 3.48 &  4.65 \\
BGRU & Word2Vec (S) &    Sequence & Att &       0.85 &    0.83 & \textbf{0.84} &      0.86 & 3.79 &  5.05 \\
\bottomrule
\end{tabular}
\caption{Evaluation results on the development subset. We compare our approach (bottom) to previous text based approaches (top). The reported results represent the \textit{best} achieved results during our experiments for a single fold (80\%) of the training data.}
\label{tab:evaluation_results}
\end{table*}

\begin{table}[htbp]
\centering
\begin{tabular}{p{1.3cm}ll|rrrr|rr}
\toprule
Embedding &  Pooling &       Data &   Pre &   Rec &    F1 & Acc &  MAE  \\
\midrule
         \multirow{2}{*}{BERT} &    \multirow{2}{*}{Mean} &     cv &       \textbf{0.82} &    0.69 & \textbf{0.69} &      \textbf{0.80} & 4.08  \\
         &     &    dev &       0.66 &    0.56 & 0.55 &      0.66 & 5.19  \\
         \hline
         \multirow{2}{*}{ELMo} &    \multirow{2}{*}{Mean} &     cv &       0.73 &    0.66 & 0.66 &      0.76 & 4.26  \\
         &     &    dev &       \textbf{0.73} &    \textbf{0.66} & \textbf{0.64} &      \textbf{0.72} & \textbf{4.71} \\
         \hline
          \multirow{2}{*}{fastText (S)} &     \multirow{2}{*}{Att} &     cv &       0.69 &    0.67 & 0.66 &      0.78 & 4.16  \\
 &      &    dev &       0.59 &    0.62 & 0.60 &      0.68 & 5.06  \\
 \hline
 \multirow{2}{*}{Word2Vec (S)} &    \multirow{2}{*}{Mean} &     cv &       0.71 &    \textbf{0.71} & 0.67 &      0.77 & \textbf{4.00}  \\
  &     &    dev &       0.61 &    0.64 & 0.61 &      0.70 & 4.95 \\
\bottomrule
\end{tabular}
\caption{Best performing pooling function across averaged 5-fold runs ($\mu_{best}$) regarding each embedding. Results on the dev set utilize the best average performing models on the CV set. Bold displays best performance on the dev/cv sets.}
\label{tab:best_avg_perf}
\end{table}

Lastly, to compare our sequence modeling approach to previous context-free and context-dependent techniques, we also provide a single, best-performing score for each method in \Cref{tab:evaluation_results}.
Please note that our best performing models are trained on a single fold of the training dataset, in line with other methods.
The the best experiment regarding F1 performance $(r_{best})$ is therefore:
\[
    r_{best} = \max_{k = 1, \ldots, 100} \left(\max_{f = 1, \ldots, 5} r_f^k(F1)\right)
\]
For the classification task on whether a patient is depressed or healthy, it is indicated that our sequence model with pretrained sentence-level text embeddings using BERT or Word2Vec, has achieved a macro F1 score of $0.84$.
Our sequence approach significantly outperforms previous context-free methods.
Specifically, our sequential approach is seen to perform equally well as the context-dependent~\cite{Williamson:2016:DDU:2988257.2988263} methods in terms of F1.
Therefore, our proposed multi-task framework approach outperforms other sequential text-based methods. 
In terms of regression, our best model using ELMo features achieves an MAE of $3.78$, being the best in its class for sequential depression modeling. 
As the results of our experiments indicated, Doc2Vec largely underperforms in terms of MAE and $F1$, compared to ELMo and BERT approaches, which we think is due to the limited training data available as well as its incapability to extract a context-dependent vector representation. 
One possible reason why general-purpose pretrained word embeddings are useful for depression detection is that clinical interview involves similar content to a normal conversation. 
Thus knowledge gained from large data pretraining can be effectively passed down to depression detection.  

As can be seen, the classification performance difference between all four utilized features is marginal.
While our proposed method does not outperform~\cite{Williamson:2016:DDU:2988257.2988263} in terms of MAE, we emphasize that the proposed approach is only requiring patients' response data to be utilized effectively.
Compared to~\cite{Haque2018}, all our proposed methods have superior (lower) MAE. 
Specifically, note that our average performance in \Cref{tab:pooling_comp} also outperforms the best-performing methods in~\cite{Haque2018,bjorn2019}.
This is likely due to our multi-task training scheme and improved data balancing pipeline, such that the model is prevented from overfitting towards depressed/healty patients.

\section{Conclusion}
\label{sec:conclusion}

This work proposed the use of multi-task modeling in conjunction with pretrained sentence embeddings, namely Word2Vec, fastText, ELMo and BERT, for text-based modeling depression. 
A BGRU model with an attention mechanism is utilized as the classifier. 
In our initial experiment we aim to investigate if pretraining on a large, unrelated dataset, is helpful to depression detection.
We observe large performance gains on both the held-out cross-validation dataset and the DAIC-WOZ development dataset.
Further, we investigate simple sentence averaging and its use for text-based depression detection.
Our experiments show that indeed sentence-level features should be preferred over word-level ones.
Lastly, we investigate four pooling functions (mean, max, time, attention) for depression detection.
Our results show that with the exception of time-pooling all other three approaches perform on average equally well.
Our highest fold-average macro F1 score goes as high as 0.69 and MAE as low as 4.00. 
In our final submission, we compare our best achieved models to other methods.
Our proposed model outperforms previous text-based detection approaches in terms of classification and PHQ-8 regression, culminating in a F1 score of $0.84$ and a MAE of $3.48$. 

Future studies can explore the model interpretations of modality fusion with speech, which can be compared with such linguistic results. 
Further conclusions can be drawn on whether a model will pay attention to different content since it has `heard' how the words are spoken.



%



\section*{Acknowledgment}

This work has been supported by the Major Program of National Social Science Foundation of China (No.18ZDA293). Experiments have been carried out on the PI supercomputer at Shanghai Jiao Tong University.

\ifCLASSOPTIONcaptionsoff
  \newpage
\fi


\bibliographystyle{IEEEtran}
\bibliography{paper}

\begin{thebibliography}{10}
\providecommand{\url}[1]{#1}
\csname url@samestyle\endcsname
\providecommand{\newblock}{\relax}
\providecommand{\bibinfo}[2]{#2}
\providecommand{\BIBentrySTDinterwordspacing}{\spaceskip=0pt\relax}
\providecommand{\BIBentryALTinterwordstretchfactor}{4}
\providecommand{\BIBentryALTinterwordspacing}{\spaceskip=\fontdimen2\font plus
\BIBentryALTinterwordstretchfactor\fontdimen3\font minus
  \fontdimen4\font\relax}
\providecommand{\BIBforeignlanguage}[2]{{%
\expandafter\ifx\csname l@#1\endcsname\relax
\typeout{** WARNING: IEEEtran.bst: No hyphenation pattern has been}%
\typeout{** loaded for the language `#1'. Using the pattern for}%
\typeout{** the default language instead.}%
\else
\language=\csname l@#1\endcsname
\fi
#2}}
\providecommand{\BIBdecl}{\relax}
\BIBdecl

\bibitem{cohn2009detecting}
J.~F. Cohn, T.~S. Kruez, I.~Matthews, Y.~Yang, M.~H. Nguyen, M.~T. Padilla,
  F.~Zhou, and F.~De~la Torre, ``Detecting depression from facial actions and
  vocal prosody,'' in \emph{2009 3rd International Conference on Affective
  Computing and Intelligent Interaction and Workshops}.\hskip 1em plus 0.5em
  minus 0.4em\relax IEEE, 2009, pp. 1--7.

\bibitem{deshpande2017depression}
M.~Deshpande and V.~Rao, ``Depression detection using emotion artificial
  intelligence,'' in \emph{2017 International Conference on Intelligent
  Sustainable Systems (ICISS)}.\hskip 1em plus 0.5em minus 0.4em\relax IEEE,
  2017, pp. 858--862.

\bibitem{yang2016decision}
L.~Yang, D.~Jiang, L.~He, E.~Pei, M.~C. Oveneke, and H.~Sahli, ``Decision tree
  based depression classification from audio video and language information,''
  in \emph{Proceedings of the 6th International Workshop on Audio/Visual
  Emotion Challenge}, 2016, pp. 89--96.

\bibitem{su2018lstm}
M.-H. Su, C.-H. Wu, K.-Y. Huang, and Q.-B. Hong, ``Lstm-based text emotion
  recognition using semantic and emotional word vectors,'' in \emph{2018 First
  Asian Conference on Affective Computing and Intelligent Interaction (ACII
  Asia)}.\hskip 1em plus 0.5em minus 0.4em\relax IEEE, 2018, pp. 1--6.

\bibitem{orabi2018deep}
A.~H. Orabi, P.~Buddhitha, M.~H. Orabi, and D.~Inkpen, ``Deep learning for
  depression detection of twitter users,'' in \emph{Proceedings of the Fifth
  Workshop on Computational Linguistics and Clinical Psychology: From Keyboard
  to Clinic}, 2018, pp. 88--97.

\bibitem{zich1990screening}
J.~M. Zich, C.~C. Attkisson, and T.~K. Greenfield, ``Screening for depression
  in primary care clinics: the ces-d and the bdi,'' \emph{The International
  Journal of Psychiatry in Medicine}, vol.~20, no.~3, pp. 259--277, 1990.

\bibitem{gilbody2007screening}
S.~Gilbody, D.~Richards, S.~Brealey, and C.~Hewitt, ``Screening for depression
  in medical settings with the patient health questionnaire (phq): a diagnostic
  meta-analysis,'' \emph{Journal of general internal medicine}, vol.~22,
  no.~11, pp. 1596--1602, 2007.

\bibitem{AlHanai2018}
\BIBentryALTinterwordspacing
T.~{Al Hanai}, M.~Ghassemi, and J.~Glass, ``Detecting depression with
  audio/text sequence modeling of interviews,'' in \emph{Proc. Interspeech
  2018}, 2018, pp. 1716--1720. [Online]. Available:
  \url{http://dx.doi.org/10.21437/Interspeech.2018-2522}
\BIBentrySTDinterwordspacing

\bibitem{Haque2018}
\BIBentryALTinterwordspacing
A.~Haque, M.~Guo, A.~S. Miner, and L.~Fei-Fei, ``{Measuring Depression Symptom
  Severity from Spoken Language and 3D Facial Expressions},'' nov 2018.
  [Online]. Available: \url{http://arxiv.org/abs/1811.08592}
\BIBentrySTDinterwordspacing

\bibitem{Williamson:2016:DDU:2988257.2988263}
\BIBentryALTinterwordspacing
J.~R. Williamson, E.~Godoy, M.~Cha, A.~Schwarzentruber, P.~Khorrami, Y.~Gwon,
  H.-T. Kung, C.~Dagli, and T.~F. Quatieri, ``Detecting depression using vocal,
  facial and semantic communication cues,'' in \emph{Proceedings of the 6th
  International Workshop on Audio/Visual Emotion Challenge}, ser. AVEC
  '16.\hskip 1em plus 0.5em minus 0.4em\relax New York, NY, USA: ACM, 2016, pp.
  11--18. [Online]. Available: \url{http://doi.acm.org/10.1145/2988257.2988263}
\BIBentrySTDinterwordspacing

\bibitem{Zhao2020}
Z.~Zhao, Z.~Bao, Z.~Zhang, J.~Deng, N.~Cummins, H.~Wang, J.~Tao, and
  B.~Schuller, ``{Automatic Assessment of Depression from Speech via a
  Hierarchical Attention Transfer Network and Attention Autoencoders},''
  \emph{IEEE Journal on Selected Topics in Signal Processing}, vol.~14, no.~2,
  pp. 423--434, feb 2020.

\bibitem{bjorn2019}
\BIBentryALTinterwordspacing
A.~Mallol-Ragolta, Z.~Zhao, L.~Stappen, N.~Cummins, and B.~W. Schuller, ``{A
  Hierarchical Attention Network-Based Approach for Depression Detection from
  Transcribed Clinical Interviews},'' in \emph{Proc. Interspeech 2019}, 2019,
  pp. 221--225. [Online]. Available:
  \url{http://dx.doi.org/10.21437/Interspeech.2019-2036}
\BIBentrySTDinterwordspacing

\bibitem{Rejaibi2020TowardsRD}
E.~Rejaibi, D.~Kadoch, K.~Bentounes, R.~Alfred, M.~Daoudi, A.~Hadid, and
  A.~Othmani, ``Towards robust deep neural networks for affect and depression
  recognition.'' \emph{arXiv: Human-Computer Interaction}, 2020.

\bibitem{Ma2016}
\BIBentryALTinterwordspacing
X.~Ma, H.~Yang, Q.~Chen, D.~Huang, and Y.~Wang, ``{DepAudioNet: An efficient
  deep model for audio based depression classification},'' in \emph{AVEC 2016 -
  Proceedings of the 6th International Workshop on Audio/Visual Emotion
  Challenge, co-located with ACM Multimedia 2016}.\hskip 1em plus 0.5em minus
  0.4em\relax New York, New York, USA: Association for Computing Machinery,
  Inc, oct 2016, pp. 35--42. [Online]. Available:
  \url{http://dl.acm.org/citation.cfm?doid=2988257.2988267}
\BIBentrySTDinterwordspacing

\bibitem{losada2020erisk}
{Losada, David E. and Crestani, Fabio and Parapar, Javier}, ``erisk 2020:
  Self-harm and depression challenges,'' in \emph{European Conference on
  Information Retrieval}.\hskip 1em plus 0.5em minus 0.4em\relax Springer,
  2020, pp. 557--563.

\bibitem{losada2017erisk}
------, ``erisk 2017: Clef lab on early risk prediction on the internet:
  experimental foundations,'' in \emph{International Conference of the
  Cross-Language Evaluation Forum for European Languages}.\hskip 1em plus 0.5em
  minus 0.4em\relax Springer, 2017, pp. 346--360.

\bibitem{coppersmith2015clpsych}
G.~Coppersmith, M.~Dredze, C.~Harman, K.~Hollingshead, and M.~Mitchell,
  ``Clpsych 2015 shared task: Depression and ptsd on twitter,'' in
  \emph{Proceedings of the 2nd Workshop on Computational Linguistics and
  Clinical Psychology: From Linguistic Signal to Clinical Reality}, 2015, pp.
  31--39.

\bibitem{Trotzek_metadata_for_early_detection}
M.~{Trotzek}, S.~{Koitka}, and C.~M. {Friedrich}, ``Utilizing neural networks
  and linguistic metadata for early detection of depression indications in text
  sequences,'' \emph{IEEE Transactions on Knowledge and Data Engineering},
  vol.~32, no.~3, pp. 588--601, 2020.

\bibitem{Trotzek_2018_word_embeddings}
\BIBentryALTinterwordspacing
M.~Trotzek, S.~Koitka, and C.~M. Friedrich, ``{Word embeddings and linguistic
  metadata at the CLEF 2018 tasks for early detection of depression and
  anorexia},'' in \emph{CEUR Workshop Proceedings}, vol. 2125, 2018. [Online].
  Available: \url{http://www.reddit.com/r/depression,}
\BIBentrySTDinterwordspacing

\bibitem{mikolov2013efficient}
\BIBentryALTinterwordspacing
T.~Mikolov, K.~Chen, G.~Corrado, and J.~Dean, ``Efficient estimation of word
  representations in vector space,'' in \emph{1st International Conference on
  Learning Representations, {ICLR} 2013, Scottsdale, Arizona, USA, May 2-4,
  2013, Workshop Track Proceedings}, Y.~Bengio and Y.~LeCun, Eds., 2013.
  [Online]. Available: \url{http://arxiv.org/abs/1301.3781}
\BIBentrySTDinterwordspacing

\bibitem{joulin2016bag}
A.~Joulin, E.~Grave, P.~Bojanowski, and T.~Mikolov, ``Bag of tricks for
  efficient text classification,'' \emph{arXiv preprint arXiv:1607.01759},
  2016.

\bibitem{bojanowski2017enriching}
P.~Bojanowski, E.~Grave, A.~Joulin, and T.~Mikolov, ``Enriching word vectors
  with subword information,'' \emph{Transactions of the Association for
  Computational Linguistics}, vol.~5, pp. 135--146, 2017.

\bibitem{pennington2014glove}
J.~Pennington, R.~Socher, and C.~Manning, ``Glove: Global vectors for word
  representation,'' in \emph{Proceedings of the 2014 conference on empirical
  methods in natural language processing (EMNLP)}, 2014, pp. 1532--1543.

\bibitem{Kong2020}
Q.~Kong, Y.~Wang, X.~Song, Y.~Cao, W.~Wang, and M.~D. Plumbley, ``{Source
  Separation with Weakly Labelled Data: an Approach to Computational Auditory
  Scene Analysis},'' in \emph{ICASSP 2020 - 2020 IEEE International Conference
  on Acoustics, Speech and Signal Processing (ICASSP)}.\hskip 1em plus 0.5em
  minus 0.4em\relax Institute of Electrical and Electronics Engineers (IEEE),
  apr 2020, pp. 101--105.

\bibitem{Kong2019a}
\BIBentryALTinterwordspacing
Q.~Kong, Y.~Cao, T.~Iqbal, Y.~Wang, W.~Wang, and M.~D. Plumbley, ``{PANNs:
  Large-Scale Pretrained Audio Neural Networks for Audio Pattern
  Recognition},'' dec 2019. [Online]. Available:
  \url{http://arxiv.org/abs/1912.10211}
\BIBentrySTDinterwordspacing

\bibitem{Chen2020}
H.~Chen, W.~Xie, A.~Vedaldi, and A.~Zisserman, ``{Vggsound: A Large-Scale
  Audio-Visual Dataset},'' in \emph{ICASSP 2020 - 2020 IEEE International
  Conference on Acoustics, Speech and Signal Processing (ICASSP)}.\hskip 1em
  plus 0.5em minus 0.4em\relax Institute of Electrical and Electronics
  Engineers (IEEE), apr 2020, pp. 721--725.

\bibitem{Hershey2017}
S.~Hershey, S.~Chaudhuri, D.~P. Ellis, J.~F. Gemmeke, A.~Jansen, R.~C. Moore,
  M.~Plakal, D.~Platt, R.~A. Saurous, B.~Seybold, M.~Slaney, R.~J. Weiss, and
  K.~Wilson, ``{CNN architectures for large-scale audio classification},'' in
  \emph{ICASSP, IEEE International Conference on Acoustics, Speech and Signal
  Processing - Proceedings}.\hskip 1em plus 0.5em minus 0.4em\relax Institute
  of Electrical and Electronics Engineers Inc., jun 2017, pp. 131--135.

\bibitem{Ringeval:2017:ARD:3133944.3133953}
\BIBentryALTinterwordspacing
F.~Ringeval, B.~Schuller, M.~Valstar, J.~Gratch, R.~Cowie, S.~Scherer,
  S.~Mozgai, N.~Cummins, M.~Schmitt, and M.~Pantic, ``Avec 2017: Real-life
  depression, and affect recognition workshop and challenge,'' in
  \emph{Proceedings of the 7th Annual Workshop on Audio/Visual Emotion
  Challenge}, ser. AVEC '17.\hskip 1em plus 0.5em minus 0.4em\relax New York,
  NY, USA: ACM, 2017, pp. 3--9. [Online]. Available:
  \url{http://doi.acm.org/10.1145/3133944.3133953}
\BIBentrySTDinterwordspacing

\bibitem{Gratch_thedistress}
\BIBentryALTinterwordspacing
J.~Gratch, R.~Artstein, G.~Lucas, G.~Stratou, S.~Scherer, A.~Nazarian, R.~Wood,
  J.~Boberg, D.~DeVault, S.~Marsella, D.~Traum, A.~Rizzo, and L.-P. Morency,
  ``The {Distress} {Analysis} {Interview} {Corpus} of human and computer
  interviews,'' in \emph{Proceedings of the {Ninth} {International}
  {Conference} on {Language} {Resources} and {Evaluation} ({LREC} 2014)}.\hskip
  1em plus 0.5em minus 0.4em\relax Reykjavik, Iceland: LREC, May 2014, pp.
  3123--3128. [Online]. Available:
  \url{http://ict.usc.edu/pubs/The%20Distress%20Analysis%20Interview%20Corpus%20of%20human%20and%20computer%20interviews.pdf}
\BIBentrySTDinterwordspacing

\bibitem{DeVault:2014:SKV:2615731.2617415}
\BIBentryALTinterwordspacing
D.~DeVault, R.~Artstein, G.~Benn, T.~Dey, E.~Fast, A.~Gainer, K.~Georgila,
  J.~Gratch, A.~Hartholt, M.~Lhommet, G.~Lucas, S.~Marsella, F.~Morbini,
  A.~Nazarian, S.~Scherer, G.~Stratou, A.~Suri, D.~Traum, R.~Wood, Y.~Xu,
  A.~Rizzo, and L.-P. Morency, ``Simsensei kiosk: A virtual human interviewer
  for healthcare decision support,'' in \emph{Proceedings of the 2014
  International Conference on Autonomous Agents and Multi-agent Systems}, ser.
  AAMAS '14.\hskip 1em plus 0.5em minus 0.4em\relax Richland, SC: International
  Foundation for Autonomous Agents and Multiagent Systems, 2014, pp.
  1061--1068. [Online]. Available:
  \url{http://dl.acm.org/citation.cfm?id=2615731.2617415}
\BIBentrySTDinterwordspacing

\bibitem{cummins2015review}
N.~Cummins, S.~Scherer, J.~Krajewski, S.~Schnieder, J.~Epps, and T.~F.
  Quatieri, ``A review of depression and suicide risk assessment using speech
  analysis,'' \emph{Speech Communication}, vol.~71, pp. 10--49, 2015.

\bibitem{Kroenke2009}
K.~Kroenke, T.~Strine, R.~Spitzer, J.~Williams, J.~Berry, and A.~Mokdad,
  ``\BIBforeignlanguage{English}{The phq-8 as a measure of current depression
  in the general population},'' \emph{\BIBforeignlanguage{English}{Journal of
  Affective Disorders}}, vol. 114, no. 1-3, pp. 163--173, 4 2009.

\bibitem{martin2017depression}
M.~Martin-Subero, K.~Kroenke, C.~Diez-Quevedo, T.~Rangil, M.~de~Antonio, R.~M.
  Morillas, M.~E. Lor{\'a}n, C.~Mateu, J.~Lupon, R.~Planas \emph{et~al.},
  ``Depression as measured by phq-9 versus clinical diagnosis as an independent
  predictor of long-term mortality in a prospective cohort of medical
  inpatients,'' \emph{Psychosomatic medicine}, vol.~79, no.~3, pp. 273--282,
  2017.

\bibitem{gui2019cooperative}
\BIBentryALTinterwordspacing
T.~Gui, L.~Zhu, Q.~Zhang, M.~Peng, X.~Zhou, K.~Ding, and Z.~Chen, ``Cooperative
  multimodal approach to depression detection in twitter,'' in \emph{The
  Thirty-Third {AAAI} Conference on Artificial Intelligence, {AAAI} 2019, The
  Thirty-First Innovative Applications of Artificial Intelligence Conference,
  {IAAI} 2019, The Ninth {AAAI} Symposium on Educational Advances in Artificial
  Intelligence, {EAAI} 2019, Honolulu, Hawaii, USA, January 27 - February 1,
  2019}.\hskip 1em plus 0.5em minus 0.4em\relax {AAAI} Press, 2019, pp.
  110--117. [Online]. Available:
  \url{https://doi.org/10.1609/aaai.v33i01.3301110}
\BIBentrySTDinterwordspacing

\bibitem{Peters:2018}
M.~E. Peters, M.~Neumann, M.~Iyyer, M.~Gardner, C.~Clark, K.~Lee, and
  L.~Zettlemoyer, ``Deep contextualized word representations,'' in \emph{Proc.
  of NAACL}, 2018.

\bibitem{devlin2018bert}
J.~Devlin, M.-W. Chang, K.~Lee, and K.~Toutanova, ``Bert: Pre-training of deep
  bidirectional transformers for language understanding,'' \emph{arXiv preprint
  arXiv:1810.04805}, 2018.

\bibitem{Yang:2016:DTB:2988257.2988269}
\BIBentryALTinterwordspacing
L.~Yang, D.~Jiang, L.~He, E.~Pei, M.~C. Oveneke, and H.~Sahli, ``Decision tree
  based depression classification from audio video and language information,''
  in \emph{Proceedings of the 6th International Workshop on Audio/Visual
  Emotion Challenge}, ser. AVEC '16.\hskip 1em plus 0.5em minus 0.4em\relax New
  York, NY, USA: ACM, 2016, pp. 89--96. [Online]. Available:
  \url{http://doi.acm.org/10.1145/2988257.2988269}
\BIBentrySTDinterwordspacing

\bibitem{Scherer:2014:DBA:2663204.2663238}
\BIBentryALTinterwordspacing
S.~Scherer, Z.~Hammal, Y.~Yang, L.-P. Morency, and J.~F. Cohn, ``Dyadic
  behavior analysis in depression severity assessment interviews,'' in
  \emph{Proceedings of the 16th International Conference on Multimodal
  Interaction}, ser. ICMI '14.\hskip 1em plus 0.5em minus 0.4em\relax New York,
  NY, USA: ACM, 2014, pp. 112--119. [Online]. Available:
  \url{http://doi.acm.org/10.1145/2663204.2663238}
\BIBentrySTDinterwordspacing

\bibitem{Lin2019}
\BIBentryALTinterwordspacing
L.~Lin, X.~Wang, H.~Liu, and Y.~Qian, ``{Specialized Decision Surface and
  Disentangled Feature for Weakly-Supervised Polyphonic Sound Event
  Detection},'' \emph{IEEE/ACM Transactions on Audio, Speech, and Language
  Processing}, pp. 1--1, may 2020. [Online]. Available:
  \url{http://arxiv.org/abs/1905.10091}
\BIBentrySTDinterwordspacing

\bibitem{Kao_rare2020}
C.~{Kao}, M.~{Sun}, W.~{Wang}, and C.~{Wang}, ``A comparison of pooling methods
  on lstm models for rare acoustic event classification,'' in \emph{ICASSP 2020
  - 2020 IEEE International Conference on Acoustics, Speech and Signal
  Processing (ICASSP)}, 2020, pp. 316--320.

\bibitem{x_vector}
D.~{Snyder}, D.~{Garcia-Romero}, G.~{Sell}, D.~{Povey}, and S.~{Khudanpur},
  ``X-vectors: Robust dnn embeddings for speaker recognition,'' in \emph{2018
  IEEE International Conference on Acoustics, Speech and Signal Processing
  (ICASSP)}, 2018, pp. 5329--5333.

\bibitem{Variani-icassp14}
\BIBentryALTinterwordspacing
E.~Variani, X.~Lei, E.~McDermott, I.~L. Moreno, and J.~Gonzalez-Dominguez,
  ``{Deep neural networks for small footprint text-dependent speaker
  verification},'' in \emph{ICASSP, IEEE International Conference on Acoustics,
  Speech and Signal Processing - Proceedings}, 2014, pp. 4052--4056. [Online].
  Available: \url{http://ieeexplore.ieee.org/document/6854363/}
\BIBentrySTDinterwordspacing

\bibitem{Wang2018}
\BIBentryALTinterwordspacing
Y.~Wang, J.~Li, and F.~Metze, ``{A Comparison of Five Multiple Instance
  Learning Pooling Functions for Sound Event Detection with Weak Labeling},''
  \emph{ICASSP, IEEE International Conference on Acoustics, Speech and Signal
  Processing - Proceedings}, vol. 2019-May, pp. 31--35, oct 2019. [Online].
  Available: \url{http://arxiv.org/abs/1810.09050}
\BIBentrySTDinterwordspacing

\bibitem{Mirsamadi2017}
S.~Mirsamadi, E.~Barsoum, and C.~Zhang, ``{Automatic Speech Emotion Recognition
  Using Recurrent Neural Networks with Local Attention},'' 2017.

\bibitem{le2014distributed}
Q.~Le and T.~Mikolov, ``Distributed representations of sentences and
  documents,'' in \emph{International conference on machine learning}, 2014,
  pp. 1188--1196.

\bibitem{rehurek_lrec}
R.~{\v R}eh{\r u}{\v r}ek and P.~Sojka,
  ``\BIBforeignlanguage{English}{{Software Framework for Topic Modelling with
  Large Corpora}},'' in \emph{\BIBforeignlanguage{English}{{Proceedings of the
  LREC 2010 Workshop on New Challenges for NLP Frameworks}}}.\hskip 1em plus
  0.5em minus 0.4em\relax Valletta, Malta: ELRA, May 2010, pp. 45--50,
  \url{http://is.muni.cz/publication/884893/en}.

\bibitem{sechidis2011stratification}
K.~Sechidis, G.~Tsoumakas, and I.~Vlahavas, ``On the stratification of
  multi-label data,'' \emph{Machine Learning and Knowledge Discovery in
  Databases}, pp. 145--158, 2011.

\bibitem{KingmaB14}
\BIBentryALTinterwordspacing
D.~P. Kingma and J.~Ba, ``Adam: {A} method for stochastic optimization,'' in
  \emph{3rd International Conference on Learning Representations, {ICLR} 2015,
  San Diego, CA, USA, May 7-9, 2015, Conference Track Proceedings}, Y.~Bengio
  and Y.~LeCun, Eds., 2015. [Online]. Available:
  \url{http://arxiv.org/abs/1412.6980}
\BIBentrySTDinterwordspacing

\bibitem{Glorot10understandingthe}
X.~Glorot and Y.~Bengio, ``Understanding the difficulty of training deep
  feedforward neural networks,'' in \emph{In Proceedings of the International
  Conference on Artificial Intelligence and Statistics (AISTATS’10). Society
  for Artificial Intelligence and Statistics}, 2010.

\bibitem{gong2017topic}
Y.~Gong and C.~Poellabauer, ``Topic modeling based multi-modal depression
  detection,'' in \emph{Proceedings of the 7th Annual Workshop on Audio/Visual
  Emotion Challenge}.\hskip 1em plus 0.5em minus 0.4em\relax ACM, 2017, pp.
  69--76.

\bibitem{zhai_chunking}
\BIBentryALTinterwordspacing
F.~Zhai, S.~Potdar, B.~Xiang, and B.~Zhou, ``Neural models for sequence
  chunking,'' \emph{CoRR}, vol. abs/1701.04027, 2017. [Online]. Available:
  \url{http://arxiv.org/abs/1701.04027}
\BIBentrySTDinterwordspacing

\end{thebibliography}
\end{document}